\documentclass[10pt,twocolumn,letterpaper]{article}

\usepackage[pagenumbers]{cvpr} 

\definecolor{cvprlightgreen}{rgb}{0.0, 0.9, 0.4}
\usepackage[pagebackref,breaklinks,colorlinks,allcolors=cvprlightgreen]{hyperref}
\usepackage{multirow}
\usepackage{comment}
\usepackage{amsmath, amssymb}
\usepackage{makecell}
\usepackage{tikz}
\usepackage{array}    
\usepackage{diagbox} 
\usepackage{makecell} 
\usepackage{pifont}
\usepackage{xcolor}

\usepackage{graphicx} 
\usepackage{algorithm}
\usepackage[noend]{algpseudocode}

\def\paperID{36302} 
\def\confName{CVPR}
\def\confYear{2026}

\title{LOOPE: Learnable Optimal Patch Order for Positional Encoders in Vision Transformers}

\author{Md Abtahi Majeed Chowdhury,
Md Rifat Ur Rahman,
Akil Ahmad Taki\\
Bangladesh University of Engineering and Technology
}

\begin{document}
\maketitle
\begin{abstract}

Transformers are inherently permutation invariant, making positional encoding (PE) fundamental to their success. While prior work has focused on designing absolute or relative positional encoders, nearly all approaches implicitly assume that the 1D ordering of image patches is fixed. Yet for higher-dimensional data such as images, mapping a 2D grid into a 1D sequence is itself a geometric operation—one that can distort locality, disrupt neighborhood structure, and ultimately limit how effectively PEs capture spatial relations. Surprisingly, the impact of the order of patches on positional representation has received little attention. We revisit this overlooked design dimension and show that patch order is not merely a preprocessing choice but a learnable degree of freedom that fundamentally shapes positional embedding quality. We propose LOOPE, a lightweight framework that learns an image-dependent ordering of patches. Starting from a locality-preserving space-filling curve and applying small context-aware refinements, LOOPE yields an interpretable and content-adaptive coordinate system on which standard sinusoidal encodings operate robustly. Across multiple ViT and ViT-based architectures, LOOPE improves performance and stabilizes PEs under challenging settings such as non-optimal frequency schedules. To complement standard benchmarks, we introduce a simple diagnostic task revealing that effective PEs can preserve far more positional information than previously reported. Our findings highlight learnable patch ordering as a powerful and largely untapped tool for improving positional encoding in vision Transformers.
\end{abstract}    
\section{Introduction}
\label{sec:intro}

Transformers have become the backbone of modern computer vision tasks—from image classification to image segmentation—yet they inherently lack explicit spatial ordering. Positional encoding (PE) is therefore essential for capturing spatial relationships. While numerous PE variants exist, effectively modeling 2D spatial structure remains challenging.

\begin{figure}[ht]
  \centering
  \begin{subfigure}{0.4\linewidth}
    \includegraphics[width=\textwidth]{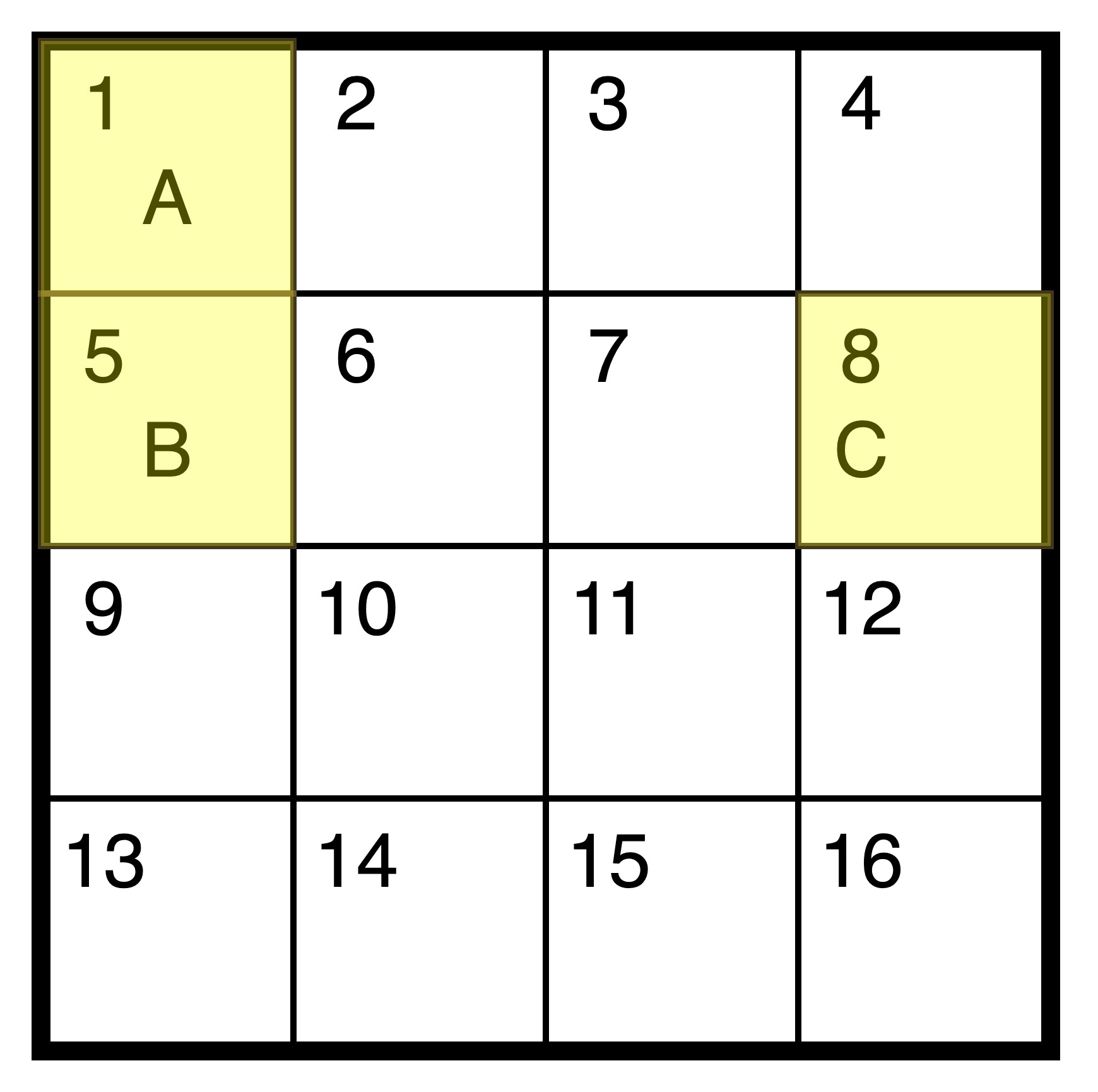}
    \caption{\(4 \times 4\)  Patched Image}
    \label{fig:image1}
  \end{subfigure}
  \begin{subfigure}{0.4\linewidth}
    \includegraphics[width=\textwidth]{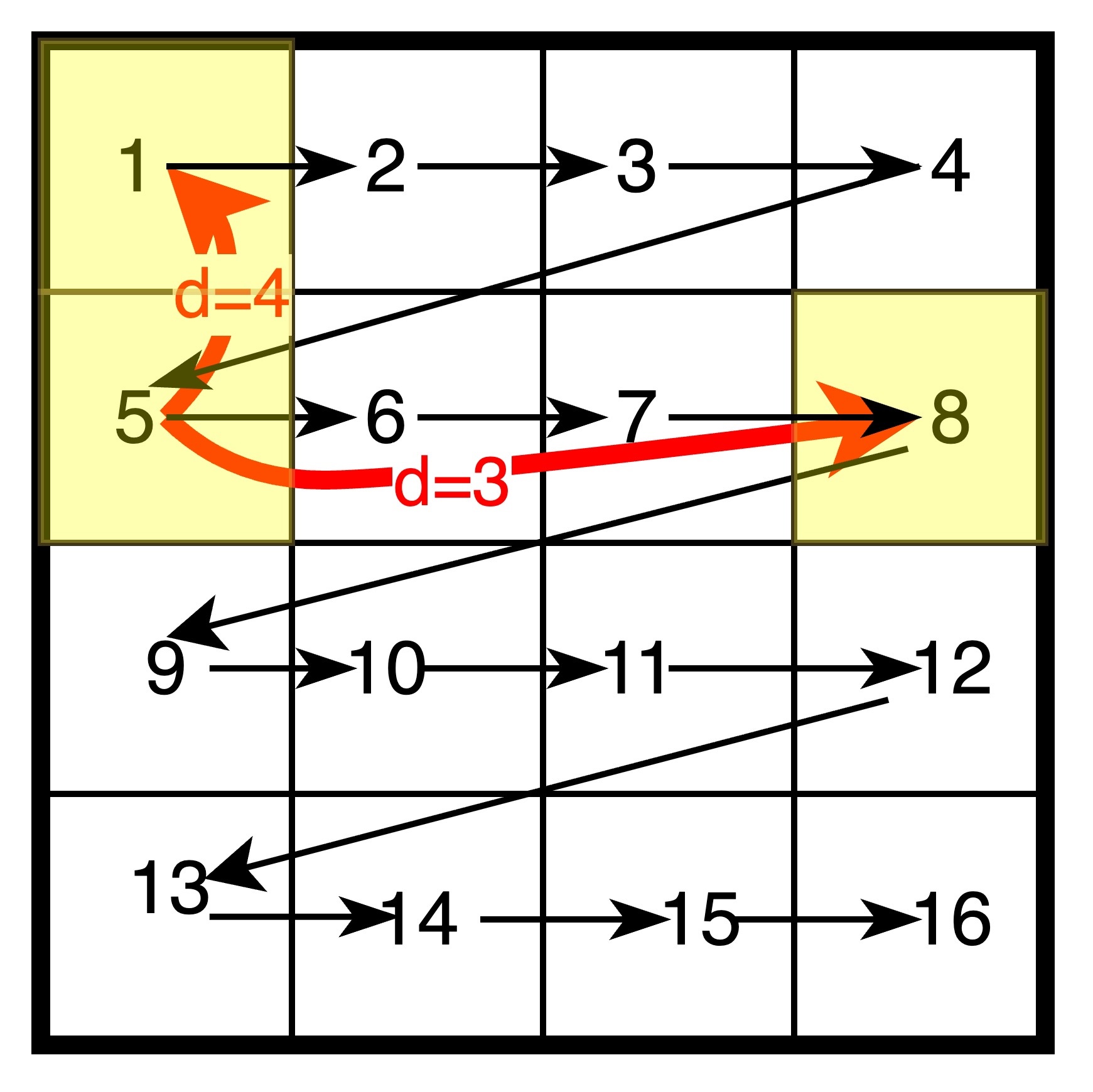}
    \caption{Typical ZigZag Order}
    \label{fig:image2}
  \end{subfigure}

  \begin{subfigure}{0.4\linewidth}
    \includegraphics[width=\textwidth]{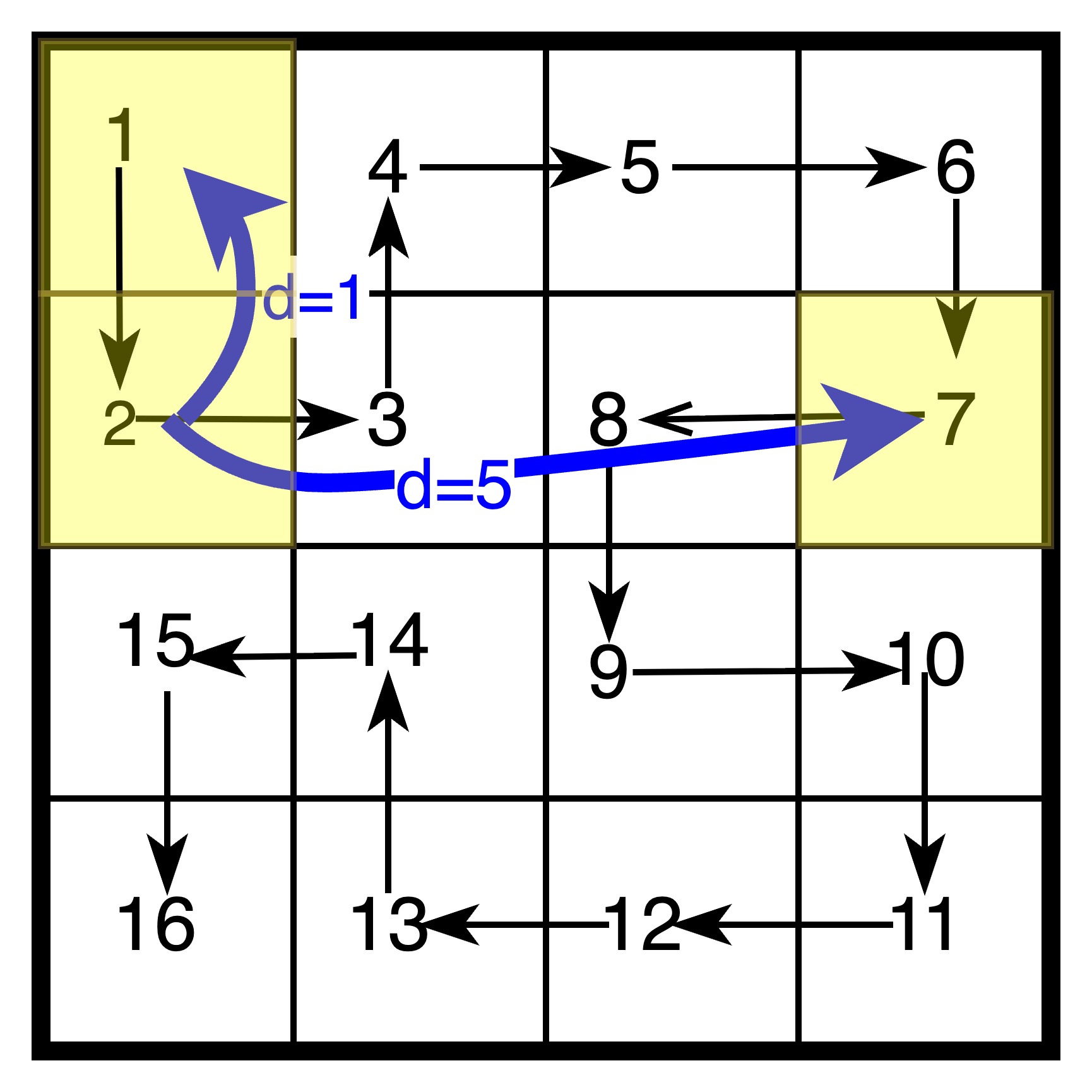} 
    \caption{Static Order}
    \label{fig:image3}
  \end{subfigure}
  \begin{subfigure}{0.4\linewidth}
    \includegraphics[width=\textwidth]{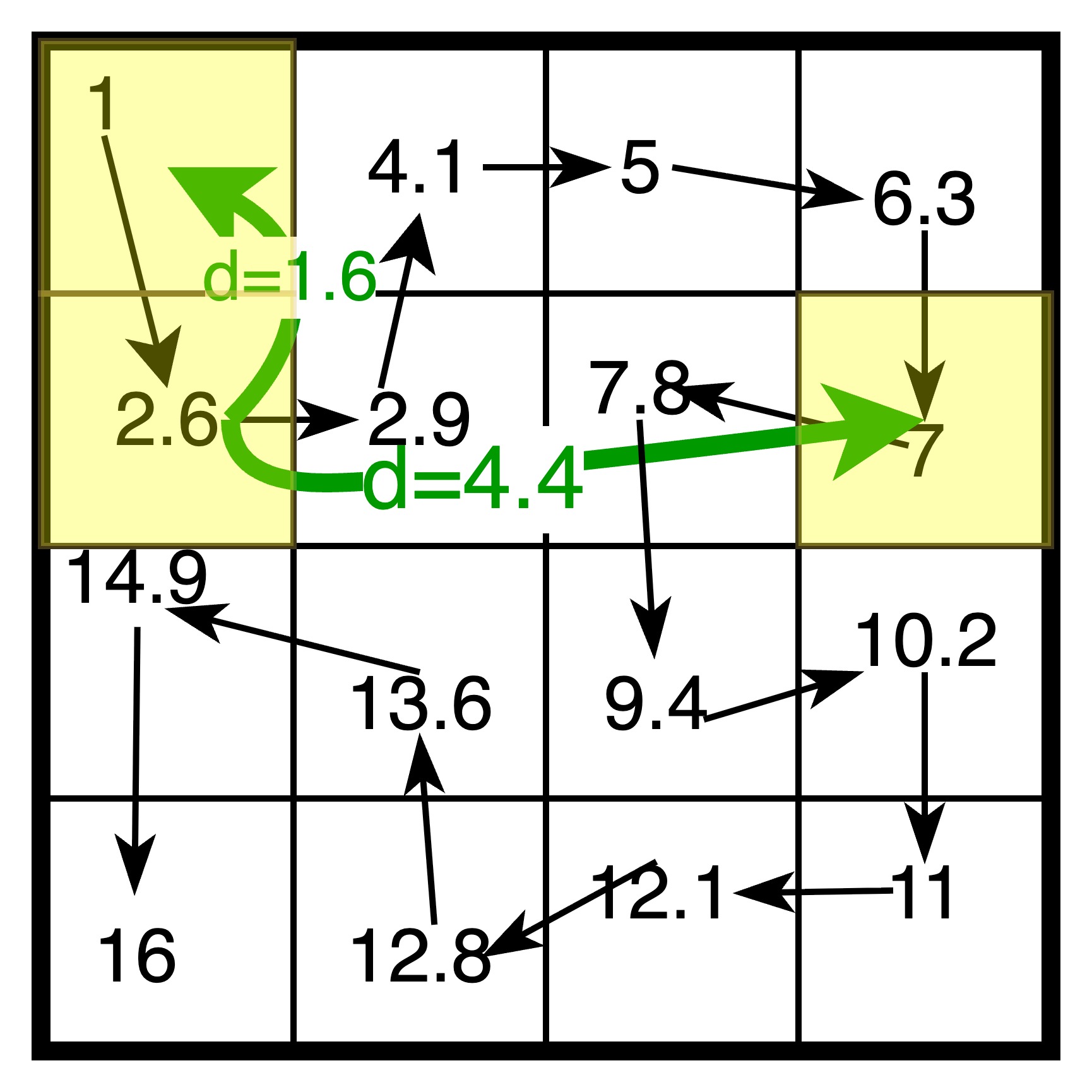}
    \caption{LOOPE}
    \label{fig:image4}
  \end{subfigure}

  \caption{Fig. \ref{fig:image2} shows that under Generic ZigZag Order, the distances between patches are \(d(B,A)=4\) and \(d(B,C)=3\), even though B is spatially and contextually closer to A. Static order (Fig. \ref{fig:image3}) improves this with \(d(B,A)=1\) and \(d(B,C)=5\), but it remains static. LOOPE (Fig. \ref{fig:image4}) dynamically reorders based on context, yielding \(d(B,A)=1.6\) and \(d(B,C)=4.4\).
}
  \label{fig:2x2grid}
\end{figure}

Traditional PEs—such as sinusoidal encodings \cite{attention}, learned absolute encodings \cite{attention}, and relative positional encodings (RPEs) \cite{selfattention}—either impose rigid spatial biases or introduce substantial parameter overhead. Conditional encoders like CPE \cite{cpvt} incorporate local information but discard global coordinates, while frequency-based approaches \cite{tancik2020fourier, rahimi2007random, xu2021positional} and Learnable Fourier Features (LFF) emphasize spectral design. However, a common assumption underlies all these methods: the 1D ordering of image patches is fixed (e.g., raster or zigzag) and independent of image content.
Specifically, we define the positional encoding as 
\begin{small}
\begin{equation}
\mathbf{PE} = \textit{Policy}(\mathbf{W} = [\omega_1, \dots, \omega_L]^T, \mathbf{X} = [x_1, \dots, x_N], \frac{\pi}{2}\delta)
\end{equation}
\end{small}
where $\omega_1,\dots,\omega_L$ denote a set of $L$ frequency values (a design hyperparameter), $x_1,\dots,x_N$ are the 1D indices assigned to each of the $N$ patches, $\delta$ is a fixed binary phase offset, and \textit{Policy} refers to the functional form of the encoder (e.g., sinusoidal or learned). The specific instantiation we adopt is given in \ref{eq:our}.

Yet none of these approaches examine \textbf{how the patch sequence itself should be constructed}. This is notable because 2D→1D flattening determines which patches become neighbors in the sequence, directly influencing locality preservation, similarity patterns, and how frequencies interact with spatial structure.

We observe that periodic PEs depend on three coupled factors: \textbf{context}, \textbf{order}, and \textbf{frequencies}. Contextual encoders \cite{rethinkingpe} and frequency-based methods, including LFF, explore portions of this space, but the joint effect of context and patch order is almost entirely unexplored. Additionally, although RPEs often perform well on downstream tasks, they encode positional information inside attention weights and do not preserve a reusable absolute coordinate system. As shown later in the Three-Cell Experiment (Sec.~\ref{sec:threecell}), when labels depend purely on geometry, well-structured absolute PEs can outperform strong RPEs, highlighting the continuing importance of absolute encodings. In this work, we address this gap by treating patch ordering as a learnable, context-aware variable and by studying its interaction with sinusoidal PEs. 
To summarize, 

\begin{enumerate}
    \item Our primary contribution is the introduction of \textbf{LOOPE}, a lightweight patch-ordering framework that combines a topology-preserving static Gilbert curve \(X_G\) with a context-aware refinement \(X_C\), yielding an image-dependent coordinate system that minimizes structural distortion and enhances robustness to frequency choices.\smallskip
    \item  We propose PESI (Positional Embeddings Structural Integrity), a new set of lightweight structural probes for analyzing how well positional embeddings maintain radial and directional monotonicity and symmetry—providing task-independent diagnostics that go beyond downstream accuracy \smallskip
    \item Finally, we propose a simple diagnostic task (Three-Cell Experiment) revealing that carefully structured absolute PEs can retain substantially richer spatial information than task-oriented RPEs.
\end{enumerate}

Extensive experiments across ViT and ViT-based architectures demonstrate that LOOPE not only improves performance but also provides a clearer geometric understanding of how positional information is encoded within transformer architectures.

\section{Related Works}
\label{sec:related works}

Positional encoding (PE) is crucial for ViTs as self-attention lacks spatial awareness. Absolute positional encodings (APEs) \cite{attention}, first introduced in NLP, were adapted for vision tasks where images are tokenized into patches. Models like ViT \cite{vit} and DeiT \cite{deit} use fixed sinusoidal or learnable encoders. APEs provide positional awareness but suffer from fixed-size constraints, limiting adaptability to varying resolutions in models such as CrossViT \cite{crossvit} and Swin \cite{swin}, and fail to capture relative spatial relationships. 

Relative positional encodings (RPEs) \cite{rethinkRPE, selfattention} encode pairwise relationships, generalizing across sequence lengths for NLP \cite{transxi} and vision tasks, e.g., axial attention \cite{axialdeep} and 2D-aware encodings \cite{2dawareem}. However, they discard absolute position information, limiting fine-grained spatial representation, particularly in object localization \cite{bottleneck}. 

Hybrid PE strategies balance absolute and relative cues. CP-RPE \cite{rethinkRPE} encodes horizontal and vertical distances separately but struggles with optimal mapping. CPE \cite{cpvt} generates encodings from local context, improving translation invariance but is hyperparameter-sensitive and lacks structured positional order. RoPE \cite{roformer, rope} preserves relative positioning via rotational invariance but misses 2D spatial hierarchies, whereas AS2DRoPE \cite{visionllama} introduces scalable 2D priors. Learned Fourier Features (LFF) \cite{fourierlearnable} further refine spatial representation by learning a spectral basis over fixed coordinates; however, LFF assumes a predetermined patch ordering and does not provide a mechanism for constructing or adapting the coordinate system itself.

ViTs are widely used for segmentation. SETR \cite{setr}, Segmenter \cite{segmenter}, and DPT \cite{dpt} use ViT backbones with learned APEs. Mask2Former \cite{mask2former} uses Swin or ResNet \cite{resnet} backbones, inheriting RPE-based design. 

\section{Methodology}

\begin{figure*}[t]
    \centering
    \includegraphics[width = 0.9\linewidth]{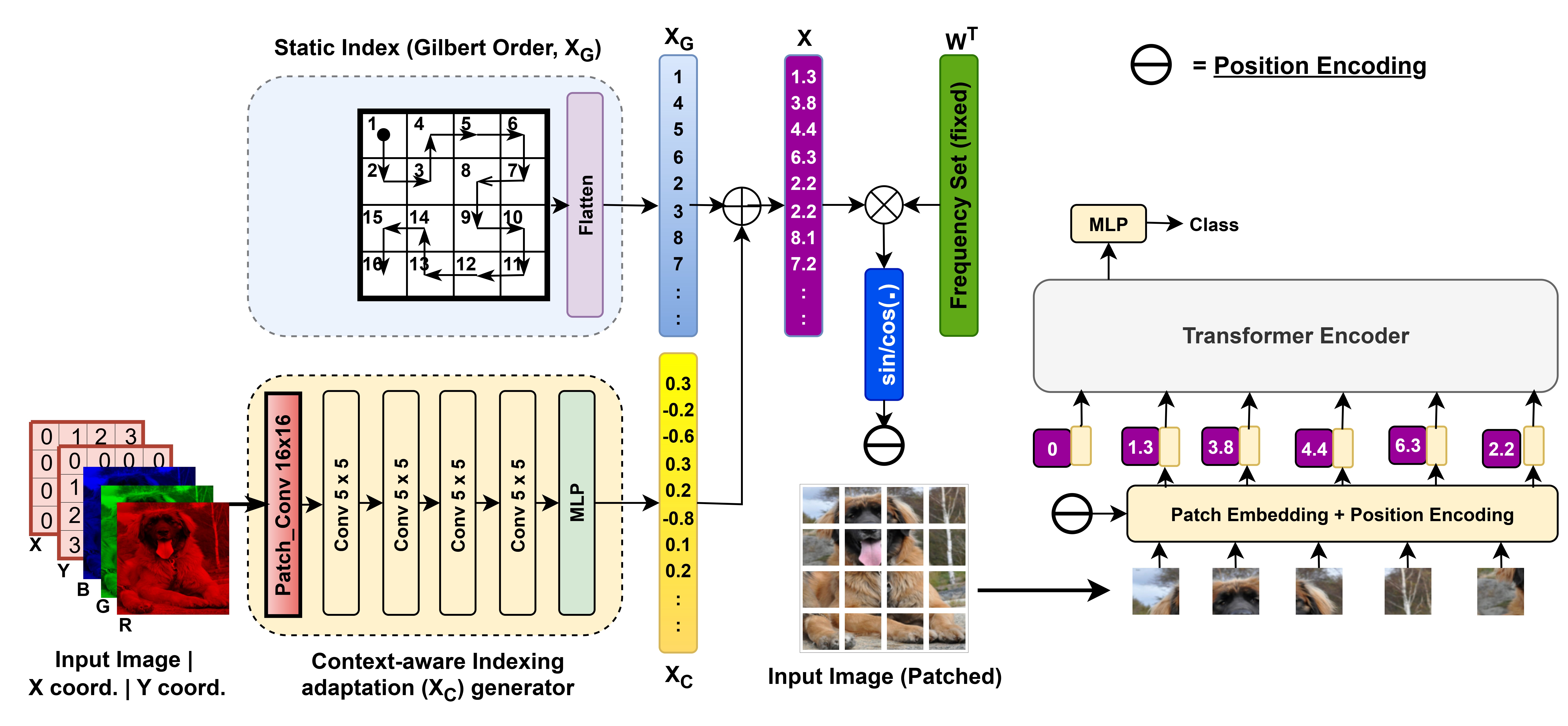}
    \caption{The complete architecture of LOOPE.}
    \label{fig:architecture}
\end{figure*}

\subsection{Motivation}

\subsubsection{A Geometric Perspective on the Problem}

Let an image be partitioned into an $h \times w$ grid of patches, giving $N=hw$ tokens. A Vision Transformer must process these tokens as a 1D sequence, which requires a bijection 
\begin{equation}
    \pi: \{1,\ldots,h\}\times \{1,\ldots,w\} \;\longrightarrow\; \{1,\ldots,N\}
\end{equation}
mapping every 2D patch coordinate to a unique sequence index. The quality of this mapping depends on how well it preserves spatial relationships: ideally, patches that are close in the grid remain close in the sequence. In geometric terms, the goal is to minimize distortion of Euclidean distances under the mapping $\pi$, i.e.\ 
\begin{equation}
\| (i,j) - (i',j') \|_2 \;\approx\; |\pi(i,j) - \pi(i',j')|
\end{equation}
so that neighborhood structure is not destroyed when flattening the image.

This is precisely the principle behind space-filling curves such as Hilbert \cite{hilbert1891} or Peano \cite{peano} traversals, which are designed to unfold two-dimensional domains into one-dimensional paths with strong locality preservation.

For ViTs, this ordering problem is not incidental but fundamental. Since self-attention is permutation-invariant, positional encodings are the only source of spatial information. If the initial 2D$\rightarrow$1D mapping $\pi$ introduces large distortions, the positional encoder must spend capacity correcting for these artifacts rather than representing meaningful structure—and even sophisticated APEs or RPEs cannot fully recover geometric relations once the coordinate system is distorted. Frequency-based methods such as LFF learn powerful positional functions but still assume a fixed ordering; our perspective instead optimizes the coordinate system itself. 

In this light, patch ordering is not a cosmetic design choice but a structural constraint that determines how much geometric information a positional encoding can ultimately preserve.

\subsubsection{Limitations of Existing Dynamic SFC Approaches}

Prior attempts to learn adaptive space-filling curves (SFCs) \cite{nsfc, esfc} highlight the importance of patch ordering, but they face two key obstacles. Most rely on discrete procedures such as Minimum Spanning Trees (MSTs) to build Hamiltonian traversals, which cannot be optimized end-to-end because they lack differentiable loss functions. GNN-based variants \cite{nsfc} estimate edge weights before applying MST heuristics, but this introduces substantial computational overhead. Moreover, these methods optimize curves independently of image content and therefore do not yield sample-specific orderings. While suitable for small graphs, such pipelines are far too heavy for positional encodings in ViTs, which must remain lightweight and efficient. In short, existing contextual SFC generators are either non-differentiable or computationally impractical—leaving open the need for a method that is both trainable and efficient in the transformer setting.

\subsection{Proposed Method}

To address these limitations, we propose \textbf{LOOPE}, a learnable patch-ordering framework that unifies two complementary components: a static space-filling order $\mathbf{X_G} \in \mathbb{Z}_{+}^{1 \times N}$ and a context-aware refinement $\mathbf{X_C} \in [-1,1]^{1 \times N}$, where $N$ is the number of patches.\textbf{ The static part ensures stability and locality preservation, while the dynamic part adapts to image content}, yielding an indexing scheme that is both efficient and trainable. In the following, we first describe the static order and then detail the context-aware refinement.

\textbf{Static Patch Index (Gilbert Order, $X_G$):} 
The Hilbert curve maps a $2^n \times 2^n$ grid, $n \in \mathbb{Z}_+$, into a 1D sequence while preserving locality, but it is restricted to square grids of power-of-two size. To generalize, we adopt the \textbf{Gilbert order} \cite{gilbert}, which recursively partitions arbitrary rectangular grids while maintaining strong spatial coherence (see Fig.~\ref{fig:image3}). This improves over typical zigzag order by reducing locality distortion, but its effectiveness still depends on the chosen frequency set, as analyzed in Table~\ref{tab:freq}.

\begin{figure}[h]
    \centering
    \includegraphics[width=1.0\linewidth]{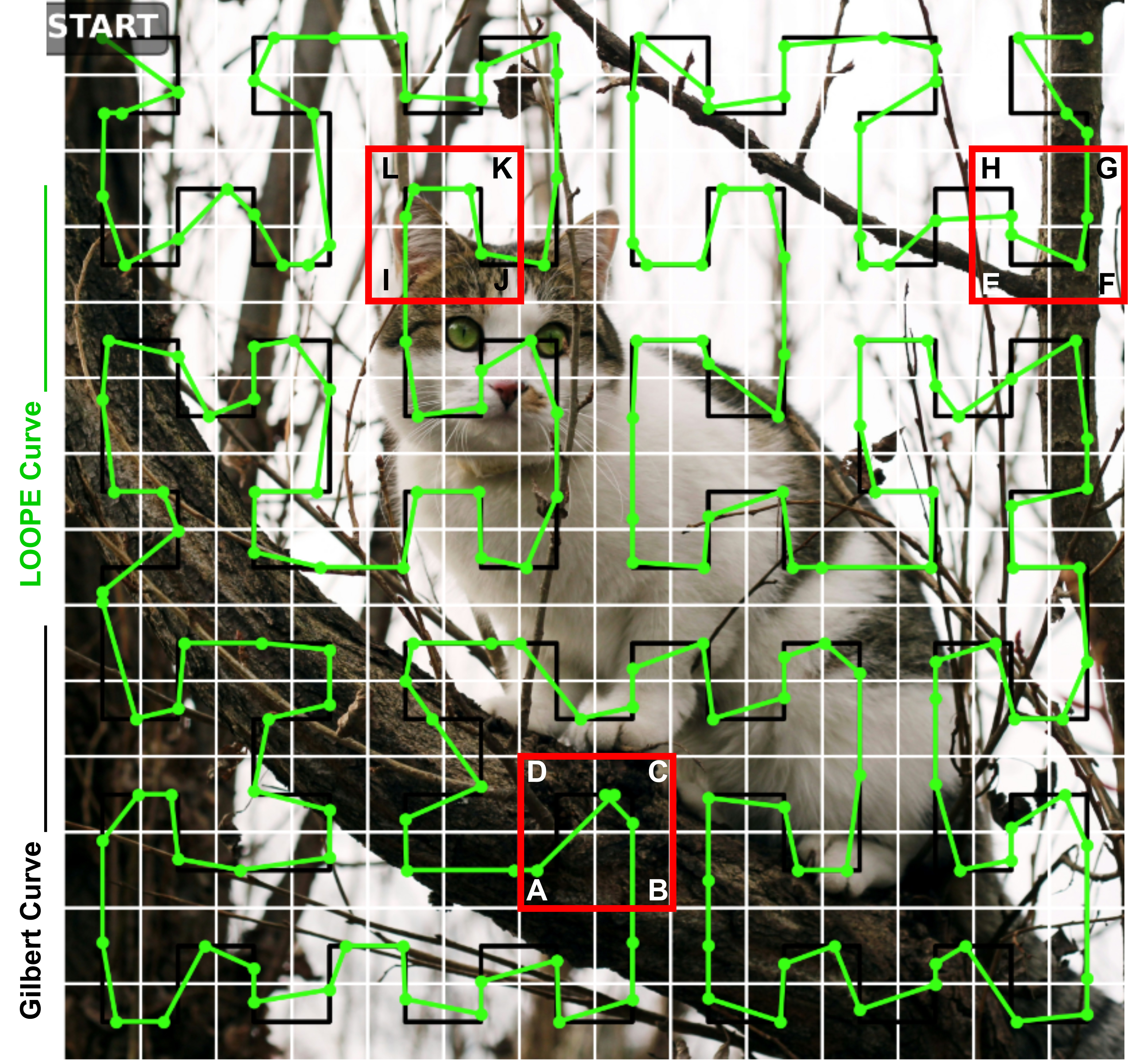}
    \caption{Here is an example of the complete generated curve. i) \(\mathbf{d_{IL}}\downarrow\) to capture ear. ii) \(\mathbf{d_{EH}}\downarrow(\text{sky to sky})\) and \(\mathbf{d_{EF}}\uparrow(\text{sky to trunk})\). iii) \(\mathbf{d_{CD}}\downarrow\) to swap C(Branch), D(Branch+Twig). Seemingly, A(Branch) is more similar to C than D, making \(A\underline{DC}B\to A\underline{CD}B\). Contextually similar patches are drawn closer. Here, \(\downarrow =\) distance decreased, \(\uparrow =\) increased.} 
    \label{fig:loope curve}
\end{figure}

\textbf{Context-Aware Index Adaptation ($X_C$):} 
The static Gilbert order $X_G$ offers stability and locality preservation, but it remains tied to the hand-crafted choice of frequency set (Table~\ref{tab:freq}). To address this, we introduce a learnable \emph{context-aware bias} $X_C$ that adapts the patch order to the input image itself. As shown in Fig.~\ref{fig:architecture}, the generator $G$ —a lightweight CNN consisting of a patch-level convolution followed by three $5 \times 5 $ convolutional layers and a final MLP (see \textit{Supp. A.2} for the full architecture) ;
 takes as input the concatenated tensor $[ \text{Image}=I_0 \in \mathbb{R}^{3\times H \times W},\, \text{coordinates}=x,y \in \mathbb{R}^{1\times H \times W}] \in \mathbb{R}^{5\times H \times W}$ and outputs continuous index offsets. These offsets are added to $X_G$, yielding a refined ordering like in Fig.~\ref{fig:loope curve},~\ref{fig:image4} that is both content-adaptive and differentiable.

This refinement introduces two decisive properties. First, it \textbf{mitigates frequency sensitivity}: unlike methods tied to a fixed frequency design, $X_C$ adaptively corrects the encoding, keeping it stable across different frequency sets (Table~\ref{tab:freq}). Second, it enables \textbf{fractional indexing}, where patch indices are no longer restricted to integers—allowing patches to be pulled closer, pushed apart, or locally reordered while preserving differentiability. This flexibility gives $X_C$ the freedom to offset poor frequency choices and to adjust neighborhood geometry in a lightweight manner (see $\partial \mathrm{PE}/\partial W$ analysis in \textit{Supp.A.3}). Thus, while $X_C$ introduces spatial awareness, its central role is to extend $X_G$ into a geometry-preserving, \emph{frequency-robust} framework—far more than a simple spatial inductive bias.

\textbf{Final Formulation:} 
Combining the static Gilbert order $X_G$ with the context-aware refinement $X_C$, LOOPE defines the final positional encoding as
\begin{equation}
    \mathbf{PE}(X) \;=\; \sin\!\Big(W(X_G + X_C) + \tfrac{\pi}{2}\,\delta \Big),
    \label{eq:our}
\end{equation}
where $W = [\omega_1,\dots,\omega_L]^T$ denotes the chosen frequency set (a design choice) and $\delta \in \mathbb{Z}_2^{L \times N}$ is a fixed phase matrix with entries $\delta_{i,j} = (i+j) \bmod 2$.

To understand why $X_C$ improves robustness to frequency choices, consider the $j$-th sinusoidal channel of Eq. \ref{eq:our}. A sensitivity analysis gives 
\begin{small}
\begin{equation}
\frac{\partial \text{PE}_j}{\partial \omega_j} = \left(X_G + X_C + \omega_j \frac{\partial X_C}{\partial \omega_j}\right) \cos\left(\omega_j(X_G + X_C) + \frac{\pi}{2}\delta_j\right)
\label{eq:freq_sensitivity}
\end{equation}
\end{small}
Crucially, $X_C$ does not take $\omega_j$ as an explicit input. Instead, its dependence on $\omega_j$ is implicit and emerges through the learning process: selecting a different frequency set \textit{prior} to training leads to a different learned refinement $X_C$. The derivative highlights two terms: a static geometric component $(X_G + X_C)$, and an adaptive correction $\omega_j \frac{\partial X_C}{\partial \omega_j}$ that captures how the network compensates for the specific initial frequency choice during training. This adaptive capacity enables LOOPE to reshape its effective positional geometry, explaining its stable performance even with non-optimal frequencies (Table \ref{tab:freq}; full derivation in Supp. A.3).

Equation~\ref{eq:our} summarizes the essence of LOOPE: a \textbf{stable static order} enriched by a \textbf{learnable, content-adaptive bias}, producing a lightweight yet flexible encoding that preserves geometry while reducing design sensitivity.

\subsection{Positional Embeddings Structural Integrity (PESI) Metrics}

Existing evaluation of positional encodings relies almost exclusively on downstream task accuracy, which conflates the PE's contribution with the backbone's learned representations. This makes it difficult to diagnose why one PE outperforms another or to predict how a PE will behave under distribution shift. We propose PESI as a set of task-independent structural probes that directly measure whether a positional embedding preserves the geometric properties one would expect from a well-structured coordinate system: (i) undirected monotonicity—similarity should decay with distance; (ii) directed monotonicity—this decay should hold consistently across all angular directions; and (iii) asymmetry—the embedding may break radial symmetry to encode directional information, and this trade-off should be quantified. 

We note that natural images do contain directional biases (e.g., dominance of cardinal orientations), so perfect symmetry is not necessarily optimal. Rather, PESI provides a vocabulary for characterizing the geometric structure of PEs and for comparing them on common ground. 

Given a positional embedding tensor $\mathbf{P} \in \mathbb{R}^{h \times w \times D}$, define the cosine similarity matrix centered at $(x, y)$: 
\begin{equation}
    E_{(x,y)}(i,j) = \frac{ P_{(x,y)} \cdot P_{(i,j)} }{ \| P_{(x,y)} \| \| P_{(i,j)} \| }.
\end{equation}

\textbf{Undirected monotonicity.}
The \textit{radial average similarity function} is:
\begin{equation} \label{mu}
\mu_{(x,y)}(r) = \frac{1}{|B_r|} \sum_{(i,j) \in B_r} E_{(x,y)}(i,j)
\end{equation}
where \( B_r \) is the set of positions at radius \( r \). We compute
\textit{Spearman’s rank correlation}:
\begin{equation}
\rho_{(x,y)} = 1 - \frac{6 \sum_{r} d_r^2}{|R|(|R|^2 - 1)} 
\end{equation}
where \( d_r \) is the rank difference between \(level(r)\) and
\(\mu_{(x,y)}(r)\), and \( |R| \) is the total number of radial levels. The
undirected monotonicity score is then
\begin{equation}
M_U = \frac{1}{hw} \sum_{x=0}^{h-1} \sum_{y=0}^{w-1} \big(1-\rho_{(x,y)} \big),
\end{equation}
where a higher \(M_U\) indicates stronger undirected monotonicity across the
grid. Ideally, it should approach 2.

\textbf{Directed monotonicity.}
We quantize \(2\pi\) into \( N = \frac{2\pi}{\delta} \) directional buckets
(with quantization angle \(\delta\)). For each cell \((i,j)\) relative to
\((x,y)\), compute:
\begin{equation}
    \theta_{(x,y)}(i,j) = \operatorname{atan2}(j-y,\, i-x)
\end{equation}
where \(\operatorname{atan2}(y,x)\) returns the angle in \((-\pi, \pi]\).
Assign the cell to bucket
\begin{equation}
k = \left\lfloor \frac{\theta_{(x,y)}(i,j)}{\delta} \right\rfloor \mod N.
\end{equation}
Within each bucket \(k\), order the cells by radial distance and compute
Spearman's rank correlation:
\begin{equation}
\rho_{(x,y)}^k = 1 - \frac{6 \sum_{r} d_r^2}{|R_k|(|R_k|^2-1)},
\end{equation}
where \(d_r\) is the rank difference between the radius \(r\) and the
corresponding similarity values in bucket \(k\), and \(|R_k|\) is the number
of elements in bucket \(k\). The mean correlation per cell is then
\begin{equation}
\bar{\rho}_{(x,y)} = \frac{1}{N}\sum_{k=0}^{N-1}\rho_{(x,y)}^k
\end{equation}
and the global directed monotonicity measure is defined as
\begin{equation}
M_D = \frac{1}{hw}\sum_{x=0}^{h-1}\sum_{y=0}^{w-1}\big(1-\bar{\rho}_{(x,y)}\big).
\end{equation}
With varying \(N\) and \(\delta\), we can investigate how precisely the
encoder maintains monotonicity along different directions. As
\(N\rightarrow 1\), \(M_D\) reduces to the undirected measure \(M_U\).
A higher \(M_D\) indicates stronger directional monotonicity; ideally it
approaches 2.

\textbf{Undirected asymmetry.}
For each center \((x,y)\), let \(B_r\) be the set of cells at radius \(r\)
from \((x,y)\). We define the standard deviation of the cosine similarity
values as
\begin{small}
\begin{align}
\sigma_{(x,y)}(r) &= \sqrt{ \frac{1}{|B_r|} \sum_{(i,j) \in B_r} \left( E_{(x,y)}(i,j) - \mu_{(x,y)}(r) \right)^2 }.
\end{align}
\end{small}
The coefficient of variation at radius \(r\) is then
\begin{equation}
\mathrm{CV}_{(x,y)}(r) = \frac{\sigma_{(x,y)}(r)}{\mu_{(x,y)}(r)}.
\end{equation}
Averaging over all radial distances \(r \in R\) yields the undirected
asymmetry measure at \((x,y)\):
\begin{equation}
A_{SU}'(x,y) = \frac{1}{|R|} \sum_{r \in R} \mathrm{CV}_{(x,y)}(r),
\end{equation}
and the global undirected asymmetry is defined as
\begin{equation}
    A_{SU} = \frac{1}{HW} \sum_{x=1}^{H} \sum_{y=1}^{W} A_{SU}'(x,y).
\end{equation}
For complete symmetry, \(|A_{SU}|\rightarrow 0\). In practice, there is no
single ideal value of undirected asymmetry, since many encoders increase
$A_{SU}$ to achieve sharper directional monotonicity $M_D$; if $A_{SU}=0$,
there is no directional information in the embedding. Detailed algorithms
are provided in \textit{Supp. A.5.1, A.5.2, A.5.3}.

\section{Experiments}

We evaluate LOOPE along four complementary axes. First, we assess classification accuracy across five ViT architectures and three standard benchmarks (Sec. \ref{sec:exp1}), followed by semantic segmentation on Cityscapes (Sec. \ref{sec:exp2}). Next, we introduce a controlled geometric probing task—the Three-Cell Experiment—that isolates how much positional information different PEs truly preserve, independent of visual content (Sec. \ref{sec:threecell}). We then apply our proposed PESI metrics to analyze the structural properties of learned embeddings (Sec. \ref{sec:exp4}). Finally, ablation studies examine frequency robustness, resolution scaling, embedding visualization (Sec. \ref{sec:ablation}).

\subsection{Experimental Setup}
\label{sec: exp_setup}
\textbf{Implementation Details.} Experiments run on a single RTX 5090 (32GB VRAM). Models are trained for 150 epochs using Adam, a cosine LR schedule ($10^{-3}$ to $2.5\times10^{-5}$), standard augmentations, and an 80-10-10 data split. Batch sizes are 96 for Oxford-IIIT and 64 for CIFAR-100, ImageNet-1k, and Three-Cell. Oxford-IIIT resolution-scaling uses $384\times384$ (batch size 32); CrossViT uses $240\times240$ with mixed $12\times12$ and $16\times16$ patches. To isolate PE contributions, all models utilize ImageNet-1K pretrained backbones (default DeiT-Base) with PEs trained from scratch. LOOPE introduces negligible computational overhead, increasing ViT-Base parameters by only 0.12\% (85.8M to 85.9M).

\textbf{Initialization and End-to-End Training.} While models are initialized with ImageNet-1K pretrained backbones to isolate PE contributions, the \textit{entire network} (backbone and PE) is subsequently trained end-to-end. To ensure LOOPE's performance gains are not merely an artifact of its ability to adapt to pretrained weights, we also evaluated models trained entirely from scratch (no ImageNet pretraining). Under this strict regime, LOOPE still maintains its distinct advantage (See Supp. B.1). This confirms that the geometric benefits of context-aware patch ordering are fundamental and agnostic to the initialization strategy.

\subsection{Experiments on Image Classification}
\label{sec:exp1}
\subsubsection{Comparison against 1-D Positional Encoders}

\begin{table}[ht]
    \centering
    \resizebox{1\linewidth}{!}{
    \begin{tabular}{ccccccc}
        \toprule
        \rotatebox{90}{\centering{Dataset}} & Model & No PE & Learnable & Sinusoid & \makecell{Static \\ \((X_G)\)} & \makecell{LOOPE\\\((X_G+X_c)\)}\\ 
        \midrule
        \multirow{5}{*}{\rotatebox{90}{Oxford-IIIT}} & ViT-Base \cite{vit}  & 83.6\% & 84.6\% & 85.3\% & 84.2\% & \textbf{88.1\%}  \\
        & DeiT-Base \cite{deit}  & 88.9\% & 89.4\% & 86.3\% & 89.0\% & \textbf{89.8\%}   \\
        & DeiT-Small \cite{deit} & 83.8\% & 83.8\% & 83.7\% & 80.6\% & \textbf{84.5\%}  \\
        & CaiT \cite{cait} & 87.4\% & 89.0\% & 90.0\% & 89.6\% & \textbf{90.5\%}  \\
        & Cross-ViT \cite{crossvit} & 88.3\% & 90.9\% & 88.0\% & 89.3\% & \textbf{91.0\%}  \\
        \midrule
        \multirow{5}{*}{\rotatebox{90}{CIFAR-100}} & ViT-Base \cite{vit}  &  79.8\% & 83.0\% & 85.2\% & 87.6\% & \textbf{88.3\%}  \\
        & DeiT-Base \cite{deit} &  82.1\% & 86.3\% & 86.6\% & 86.9\% & \textbf{87.1\%} \\
        & DeiT-Small \cite{deit}  &  68.6\% & 81.6\% & 71.9\% & 77.7\% & \textbf{82.0\%}  \\
        & CaiT \cite{cait} &  77.3\% & 82.5\% & 82.3\% & 82.5\% & \textbf{83.1\%}  \\
        & Cross-ViT \cite{crossvit} &  80.5\% & 84.6\% & 86.3\% & 85.3\% & \textbf{86.8\%}  \\
        \midrule
        \multirow{5}{*}{\rotatebox{90}{ImageNet-1k}} & ViT-Base \cite{vit}  &  82.7\% & 83.6\% & 84.4\% & 84.5\% & \textbf{84.7\%}  \\
        & DeiT-Base \cite{deit} &  81.7\% & 82.1\% & 82.0\% & 83.1\% & \textbf{83.5\%} \\
        & DeiT-Small \cite{deit}  &  80.5\% & 82.2\% & 82.8\% & 82.8\% & \textbf{83.0\%}  \\
        & CaiT \cite{cait} &  80.3\% & 81.6\% & 81.6\% & 82.8\% & \textbf{83.1\%}  \\
        & Cross-ViT \cite{crossvit} &  82.4\% & 83.5\% & 83.1\% & 84.5\% & \textbf{84.8\%} \\
        \bottomrule
    \end{tabular}
    }
    \caption{Accuracy comparison of different ViT models with various PEs across Oxford-IIIT, CIFAR-100 and ImageNet-1k datasets.}
    \label{tab:Table 1}
\end{table}

From Table \ref{tab:Table 1}, LOOPE achieves the highest accuracy across all models on Oxford-IIIT, most notably with ViT, indicating enhanced fine-grained feature learning. CIFAR-100 presents greater inter-class variability, still our PE outperforms all other PEs, especially with DeiT-Small. Finally, for ImageNet-1k, LOOPE again surpasses all PEs across all models. These results demonstrate that it effectively balances structured spatial encoding with learnable adaptability, making it a robust solution.
\subsubsection{More comparisons with Positional Encoders}

\begin{table}[ht]
    \centering
    \resizebox{.9\linewidth}{!}{
    \begin{tabular}{cccc}
        \toprule
        PE & ImageNet-1k & Oxford-IIIT & CIFAR-100  \\ 
        \midrule
        CPE \cite{cpvt} & 82.7\% & 83.9\% & 79.1\% \\
        RPE \cite{rethinkRPE}  &  80.3\% &  80.5\% & 79.2\%\\
        LFF(Fourier)\cite{fourierlearnable} & 83.4\% & \textbf{90.5\%} & \textbf{89.1\%} \\
        2D Sinusoid \cite{2dtpe} \ & 83.2\% & 80.1\% & 86.3\% \\
        AS2DRoPE \cite{visionllama} &  82.9\% & 88.7\% & 86.4\%  \\
        Static \((X_G)\) \cite{hilbert1891} &  83.1\% & 89.0\% & 86.9\%  \\
        \midrule
        LOOPE \((X_G+X_c)\) & \textbf{83.5\%} & 89.8\% & 87.1\% \\
        \bottomrule
    \end{tabular}}
    \caption{Accuracy comparison of LOOPE against other advanced PEs on DeiT-Base across Oxford-IIIT, CIFAR-100 and ImageNet-1k datasets}
    \label{tab:Table 2}
\end{table}
Table \ref{tab:Table 2} compares LOOPE against advanced PEs. Fourier PE achieves the highest accuracy for Oxford IIT and CIFAR-100, due to its rich frequency encoding properties. But, LOOPE outperforms Fourier in ImageNet-1k. These results validate that LOOPE remains a strong alternative for vision tasks.

\subsection{Experiments on Semantic Segmentation}
\label{sec:exp2}
\begin{table}[ht]
    \centering
    \resizebox{1\linewidth}{!}{
    \begin{tabular}{llccccc}
        \toprule
        \rotatebox{90}{Model} & PE & Acc & mIoU & mDice & \makecell{Boundary\\F1} & MCC \\ 
        \midrule
        \multirow{7}{*}{\rotatebox{90}{SETR \cite{setr}}}
        & No PE & 80.84\% & 79.56\% & 84.53\% & 3.48\% & 88.75\% \\
        & 1D Sinusoid \cite{selfattention} & 80.96\% & 78.33\% & 82.51\% & 3.55\% & \textbf{91.99\%} \\
        & Learnable \cite{selfattention} & 81.35\% & 74.41\% & 79.26\% & 3.88\% & 88.87\% \\
        & CPE \cite{cpvt} & 80.67\% & 78.42\% & 82.71\% & 3.31\% & 87.26\% \\
        & LFF(Fourier) \cite{fourierlearnable} & 82.69\% & 80.41\% & 85.20\% & \textbf{7.83\%} & 85.64\% \\
        & 2D Sinusoid \cite{2dtpe} & 80.74\% & 78.26\% & 82.46\% & 3.04\% & 91.88\% \\
        \cmidrule(lr){2-7}
        & LOOPE & \textbf{83.63\%} & \textbf{81.37\%} & \textbf{86.17\%} & 7.44\% & 89.56\% \\
        \midrule
        \multirow{7}{*}{\rotatebox{90}{Segmenter \cite{segmenter}}}
        & No PE & 79.55\% & 79.13\% & 84.48\% & 6.23\% & 83.34\% \\
        & 1D Sinusoid \cite{selfattention} & 81.82\% & 80.59\% & 85.79\% & 18.37\% & 84.74\% \\
        & Learnable \cite{selfattention} & 81.48\% & 80.46\% & 85.73\% & 15.33\% & 84.68\% \\
        & CPE \cite{cpvt} & 79.19\% & 78.71\% & 84.00\% & 8.80\% & 82.89\% \\
        & LFF(Fourier) \cite{fourierlearnable} & 83.20\% & 81.41\% & 86.46\% & 26.60\% & 85.51\%\ \\
        & 2D Sinusoid \cite{2dtpe} & 80.93\% & 79.90\% & 85.13\% & 9.58\% & 84.08\% \\
        \cmidrule(lr){2-7}
        & LOOPE & \textbf{84.05\%} & \textbf{82.26\%} & \textbf{87.31\%} & \textbf{32.21\%} & \textbf{86.37\%} \\
        \bottomrule
    \end{tabular}
    }
    \caption{Performance comparison of PEs with SETR and Segmenter models on the Cityscapes dataset.}
    \label{tab:table 3}
\end{table}

To test the impact of LOOPE on downstream tasks, We used the Cityscapes \cite{Cityscapes} dataset for its strong sensitivity to positional information. SETR \cite{setr} and Segmenter \cite{segmenter} were chosen as base models for their state-of-the-art performance and exclusive use of ViT backbones, making them PE-sensitive. We excluded other SOTA models like Mask2Former \cite{mask2former} due to extra attention modules and custom loss, and SAM-2 \cite{sam2} for its prompt encoder with cross-attention. All experiments were conducted with a batch size of 64 and trained for 100 epochs. We used a learning rate of 0.0001 and an 80-20 train-validation split.

From Table \ref{tab:table 3}, we observe that LOOPE outperforms all PEs by learning patch order, with LFF (Fourier) closely behind — surpassing LOOPE only once in Boundary F1 for SETR — and 1D Sinusoid outperforming LOOPE in MCC.

\subsection{Three-Cell Experiment: Positional Probing Task}
\label{sec:threecell}

\begin{figure}[t]
  \centering
  \begin{subfigure}{0.4\linewidth}
    \includegraphics[width=\linewidth]{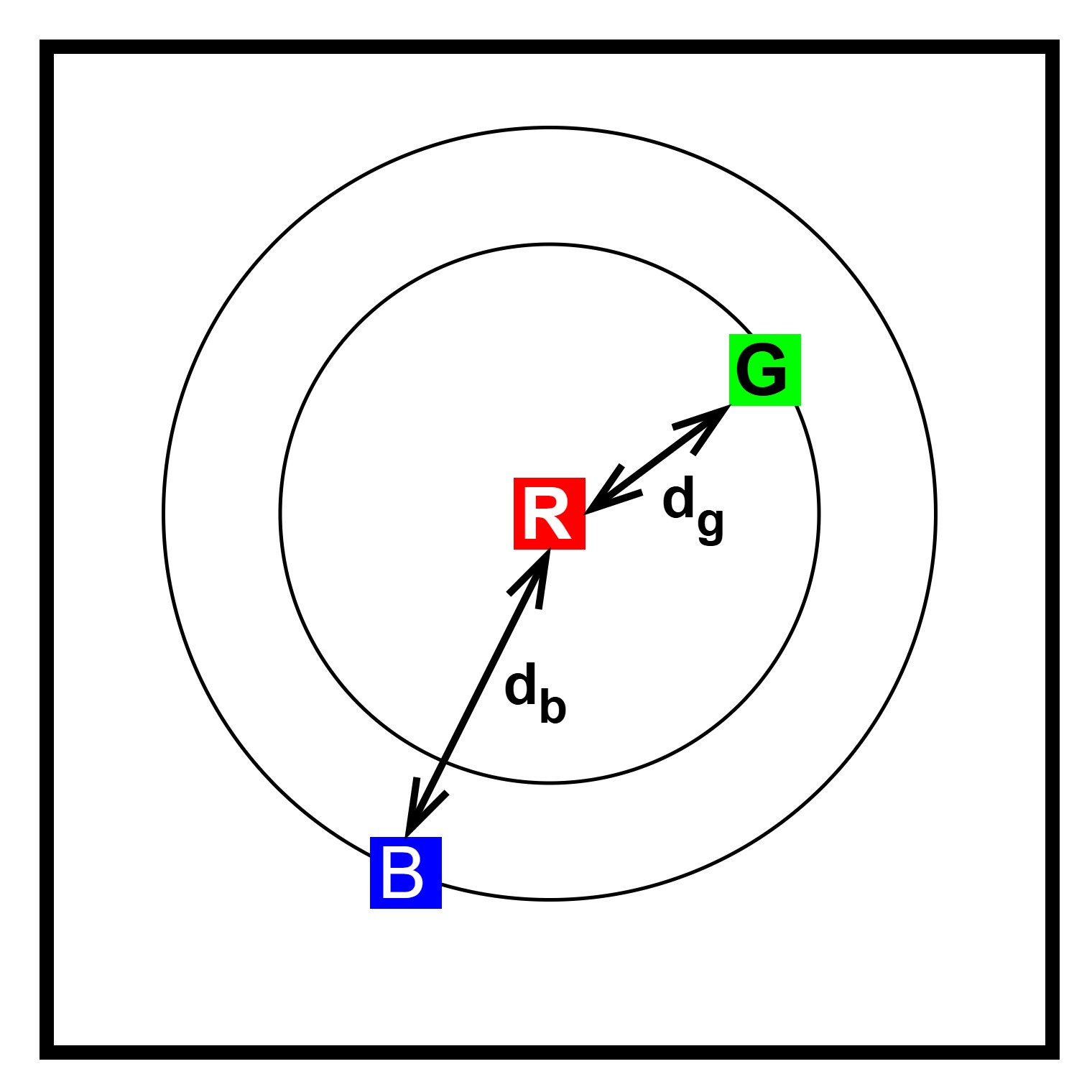}
    \caption{}
    \label{fig:threecell_distance}
  \end{subfigure}
  \begin{subfigure}{0.4\linewidth}
    \includegraphics[width=\linewidth]{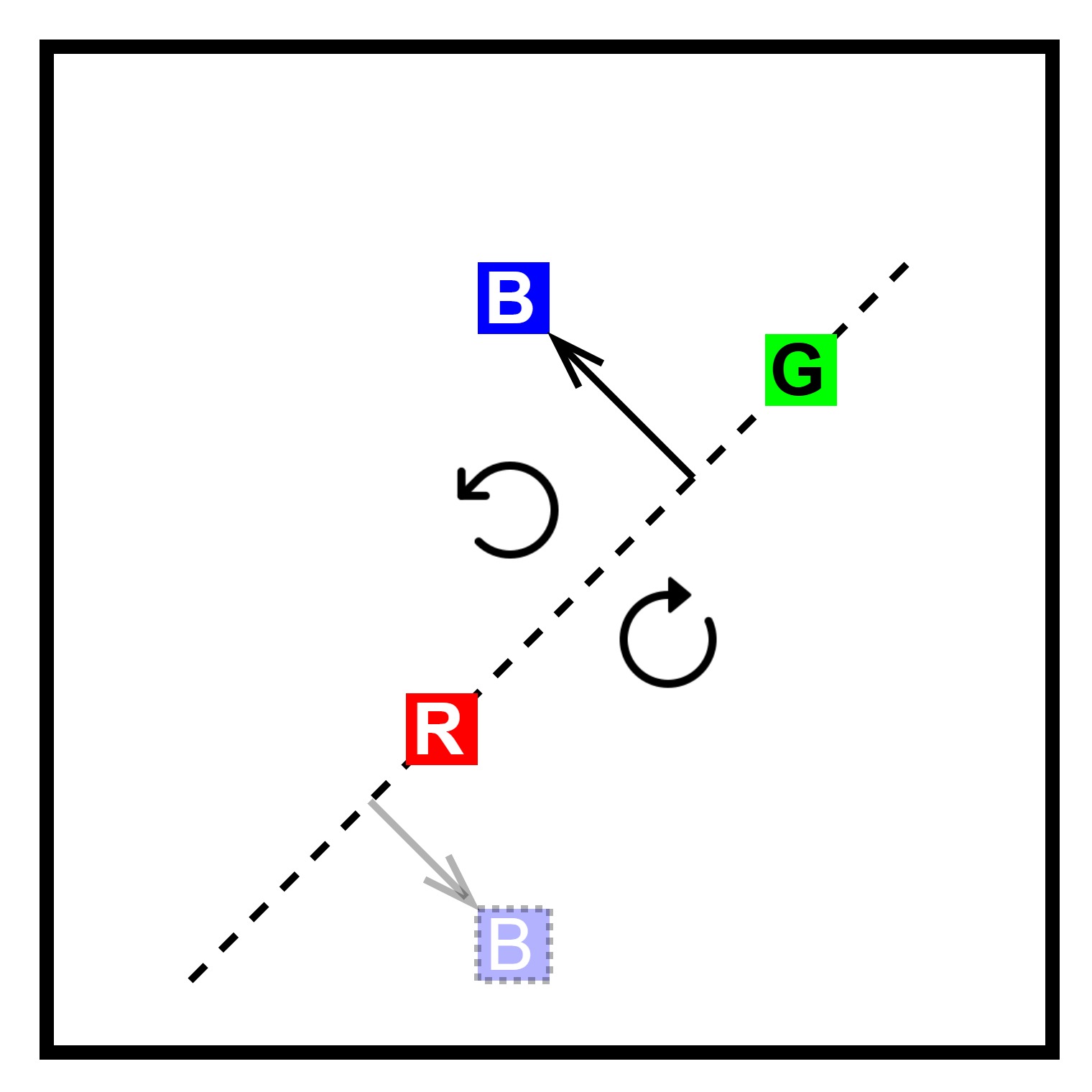}
    \caption{}
    \label{fig:threecell_orientation}
  \end{subfigure}
  \vspace{4pt}
  \begin{subfigure}{0.4\linewidth}
    \includegraphics[width=\linewidth]{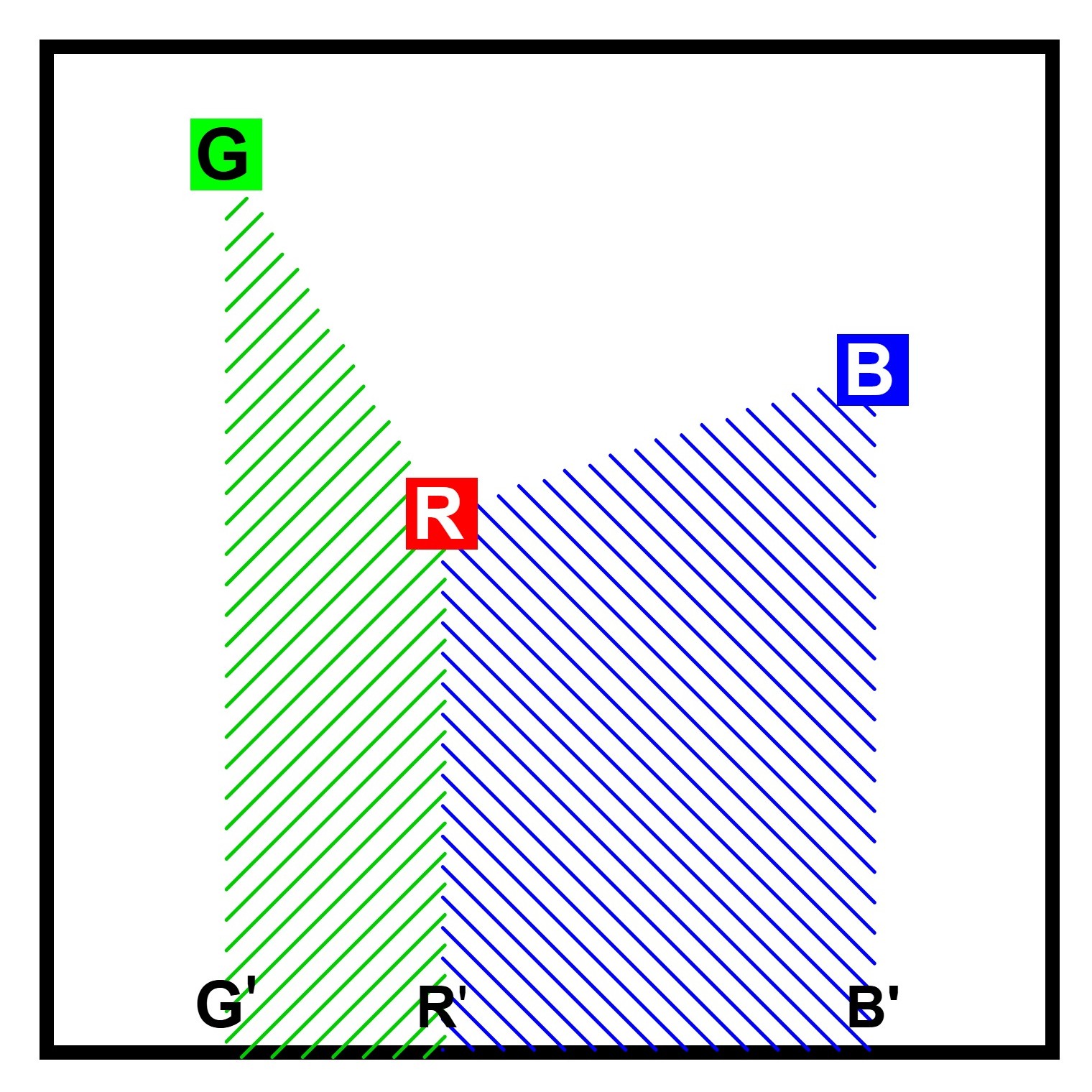}
    \caption{}
    \label{fig:threecell_area}
  \end{subfigure}
  \begin{subfigure}{0.4\linewidth}
    \includegraphics[width=\linewidth]{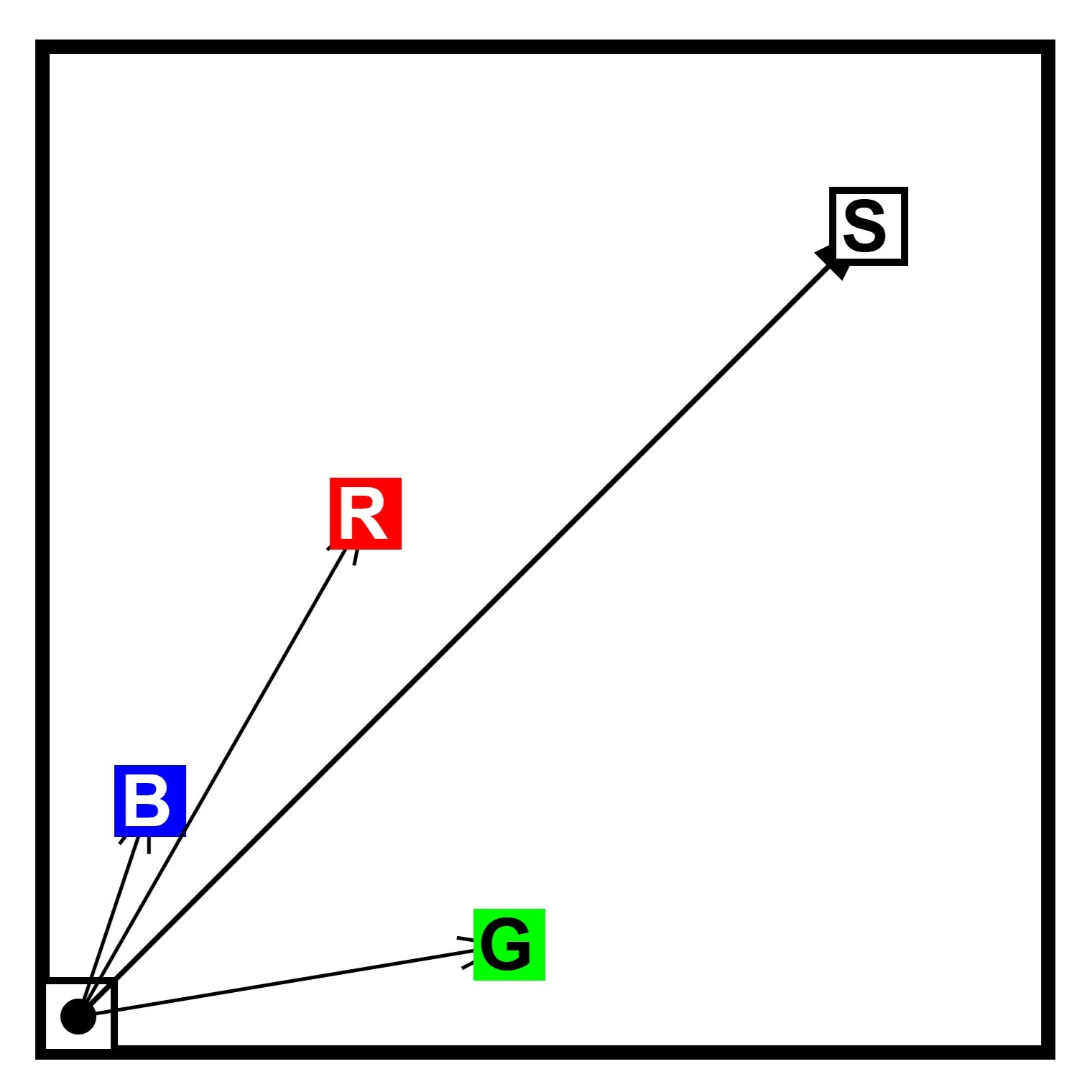}
    \caption{}
    \label{fig:threecell_vector}
  \end{subfigure}
  \caption{Three-Cell probing task. Given three colored patches (R,G,B), the
  model must: (a) decide whether $d_{RG}$ or $d_{RB}$ is larger; (b) predict
  the orientation of triangle RGB (clockwise vs.\ counterclockwise);
  (c) compare ``shadow areas'' under lines RG and RB; (d) decide whether the
  sum vector $OR + OG + OB$ lies outside the grid. Labels depend purely on
  the coordinates of R, G, and B.}
  \label{fig:three-pixel-experiment}
\end{figure}

Standard vision benchmarks often show only modest gains from positional
encodings (PEs), because natural images exhibit strong correlations between
neighboring patches and transformers can infer rough spatial structure from
content alone. This makes it difficult to isolate how much positional
information different encoders truly preserve, and in particular to compare
absolute PEs (APEs) with task-oriented relative PEs (RPEs). To obtain a more
controlled view, we design a synthetic probing task where labels depend purely
on geometry and cannot be solved from appearance.
\vspace{-1.2em}
\paragraph{Setup.} Each $224\times224$ image is divided into a $14\times14$ grid of
$16\times16$ patches. Three patches are randomly selected and colored red,
green, and blue at coordinates $(x_r,y_r),(x_g,y_g),(x_b,y_b)$, while all
other patches contain color black. The model receives only
these images and must answer geometric questions about the positions of the
three colored cells. We consider four query types
(Fig.~\ref{fig:three-pixel-experiment}) and cast them jointly as a 6-way
classification problem. Full formulas and the sampling procedure are provided
in \textit{Supp.~A.2}.

\vspace{-1.2em}
\paragraph{Geometric queries.}
The four query types are designed to probe complementary aspects of positional
structure:
(i) \textit{distance comparison} between $d_{RG}$ and $d_{RB}$, testing
undirected monotonicity;
(ii) \textit{orientation} (clockwise vs.\ counterclockwise) of the RGB
triangle, testing signed relational structure;
(iii) \textit{shadow area comparison} under RG vs.\ RB, which requires
combining offsets via addition; and
(iv) \textit{vector sum}, predicting whether
$\mathbf{p}_r + \mathbf{p}_g + \mathbf{p}_b$ leaves the $14\times14$ grid,
explicitly testing additive composition of coordinates. While the first two
can in principle be derived from relative differences, the latter two are
fundamentally difficult to reconstruct from purely RPE-based formulations.

\begin{table}[ht]
    \centering
    \resizebox{1.0\linewidth}{!}{
    \begin{tabular}{cccccc}
        \toprule
        & \multicolumn{2}{c}{\textbf{Relative Spatial Reasoning}} & \multicolumn{2}{c}{\textbf{Absolute Spatial Reasoning}} & \\
        \cmidrule(lr){2-3} \cmidrule(lr){4-5}
        \textbf{Models} & Distance & Orientation & Area & \makecell{Vector Sum} & Average \\
        \toprule
        ResNet-50\cite{resnet} & 90.8\% & 85.6\% & 89.7\% & 92.9\% & 89.8\% \\
        Inception-V3\cite{inception} & 92.8\% & 96.1\% & 93.7\% & 95.3\% & 94.5\% \\
        \toprule
        No PE &  61.9\% & 53.1\% & 60.1\% & 62.6\% & 59.4\%  \\
        Learnable \cite{attention} &  85.6\% & 89.3\% & 84.0\% & 92.2\% & 87.8\%  \\
        1D Sinusoid \cite{attention} &  90.9\% & 94.9\% & 91.3\% & \textbf{95.3\%} & 93.1\%  \\
        CPE \cite{cpvt} &  72.8\% & 63.2\% & 72.6\% & 75.0\% & 70.9\%  \\
        RPE \cite{rethinkRPE} &  73.7\% & 62.6\% & 71.0\% & 73.9\% & 70.3\%  \\
        LFF(Fourier) \cite{fourierlearnable} & 93.1\% & \textbf{96.7}\% & 92.4\% & 93.9\% & 94.0\% \\
        2D Sinusoid \cite{fourierlearnable} & 83.6\% & 60.0\% & 72.6\% & 92.3\% & 77.1\% \\
        Static \((X_G)\) \cite{hilbert1891} &  88.8\% & 95.0\% & 89.9\% & 93.8\% & 91.9\%  \\
        \toprule
        LOOPE \((X_G+X_c)\) & \textbf{93.4\%} & 95.8\% & \textbf{93.3\%} & 94.6\% & \textbf{94.3\%}  \\
        \bottomrule
    \end{tabular}
    }
    \caption{Accuracy comparison of baseline CNN models and PEs with DeiT-Base model on the Three-Cell Experiment dataset.}
    \label{tab:Table 4}
\end{table}
\vspace{-1.3em}
\paragraph{Findings.}
Table~\ref{tab:Table 4} shows three consistent trends. First, RPEs clearly
outperform the no-PE baseline but remain close to chance on the area and
vector-sum tasks, confirming that they struggle to encode additive coordinate
structure. Second, APEs (sinusoidal, LFF) achieve much higher accuracy across
all queries, indicating that absolute coordinates remain crucial even in
modern ViTs. Third, LOOPE yields the highest geometric fidelity, suggesting
that learning a content-adaptive patch ordering further improves how absolute
PEs expose positional information to the transformer.

Overall, this probing task supports our central claim: rather than relying
solely on task-oriented RPEs, developing stronger APEs—and in particular
learnable patch orderings such as LOOPE—remains a promising direction for
positional encoding research. (See Supp. B.2, B.3 for futher insights)

\subsection{Positional Embedding Structural Integrity (PESI) Metrics}
\label{sec:exp4}
\begin{table}[ht]
    \centering
    \resizebox{0.95\linewidth}{!}{
    \begin{tabular}{cccc}
        \toprule
        PE & \multicolumn{1}{c}{Undirected} & \multicolumn{1}{c}{Directed} & \multicolumn{1}{c}{Undirected} \\ 
        & Monotonicity & Monotonicity & Asymmetry \\
        & \(M_U\uparrow\) & \(M_D\uparrow\) & \(A_{SU}\) \\
        \midrule
        Learnable \cite{attention} & 1.7493 & 1.2003 & -0.7272 \\
        1D Sinusoid \cite{attention} & 1.9567 & 1.4905 & 0.1243 \\
        LFF(Fourier) \cite{fourierlearnable} & 1.9623 & \textbf{1.5230} & 0.2683 \\
        Static \((X_G)\) \cite{hilbert1891} & 1.9670 & 1.2897 & 0.0945 \\
        \toprule
        LOOPE \((X_G+X_c)\) & \textbf{1.9674} & 1.2900 & \textbf{0.0939} \\
        \bottomrule
    \end{tabular}
    }
    \caption{Comparison of PEs in terms of Undirected Monotonicity, Directed Monotonicity, and Undirected Asymmetry.}
    \label{tab: Table 5}
\end{table}

Table \ref{tab: Table 5} presents the positional fidelity indices for various APEs. For calculating directed monotonicity, the number of buckets, N is set to 60 testing. So, the \(\delta = 6^o\)
The results indicate that LOOPE achieves the highest values in both \textbf{Undirected Monotonicity} and \textbf{Undirected Asymmetry}, demonstrating its robustness. Conversely, Learnable APE performs the worst across all three metrics, indicating that its embeddings are not highly monotone. A notable observation is the asymmetry value of Learnable, which is -0.72. This negative value arises because the average cosine similarity across all cells is predominantly negative, leading to an overall asymmetry value below zero. Meanwhile, Fourier exhibits strong \textbf{Directed Monotonicity} with stable results in the undirected setting. However, it compromises radial symmetry, meaning that values on a single radius show greater instability compared to other periodic APEs. In contrast, LOOPE demonstrates the most stable radial symmetry, reinforcing its reliability in positional encoding. 

\subsection{Ablation Studies}
\label{sec:ablation}
\subsubsection{Robustness across varying Frequency Set}

\begin{table}[ht]
\centering
\resizebox{1.0\linewidth}{!}{
    \begin{tabular}{ccccc}
        \toprule
        \diagbox{Freq.}{PE} & \(\omega(i)\)  & 1D Sinusoid & \makecell{Static \\\((X_G)\)} & \makecell{LOOPE\\ \((X_G+X_c)\)} \\
        \midrule
        \makecell{\textbf{Original}} & \(0.978^i\)
        & 85.3\% & 84.6\% & \textbf{88.1\%} \\
        \midrule
        \makecell{\textbf{Arithmetic}} & \(1 - \frac{i(1 - \lambda)}{L - 1}\)
        & \underline{73.6\%} & \underline{76.5\%} & \textbf{\underline{86.6\%}} \\
        \midrule
        \makecell{\textbf{Geometric}} & \(r^i,\ r=0.9\)
        & 89.8\% & 87.3\% & \textbf{89.9\%} \\
        \midrule
        \makecell{\textbf{Random}} & \(\text{Uniform}\left(\lambda, 1\right)\)
        & 83.5\% & 87.5\% & \textbf{88.0\%} \\
        \bottomrule
        \makecell{\textbf{Sensitivity to Freq. \(\downarrow\)}} & 
        & 0.027760 & 0.022516 & \textbf{0.005244} \\
        \bottomrule
    \end{tabular}}
\caption{Robustness of LOOPE across different frequency sequences (\(\lambda = 0.0001\), \(L=768\), \(i\in1,\dots,L\)) with ViT-Base as backbone on Oxford-IIIT Dataset. For Sensitivity computation, refer to \textit{Supp.A.6.} }
\label{tab:freq}
\end{table}

The results in Table~\ref{tab:freq} demonstrate that \textbf{LOOPE} consistently outperforms other PEs, even when using a non-optimal frequency set. Notably, there is no theoretically proven optimal frequency configuration. For instance, a simple geometric sequence achieves better accuracy than the original frequency set across all PEs. \textbf{This analysis clearly highlights the robustness of LOOPE—particularly in the case of the arithmetic sequence, where accuracy drops to 73.6\%, yet LOOPE alone helps maintain it at 86.6\%.} This experiment indicates that \textbf{LOOPE} effectively mitigates frequency selection bias.

\subsubsection{Impact of LOOPE on Variable Resolution}
\begin{table}[H]
    \centering
    \resizebox{0.85\linewidth}{!}{
    \begin{tabular}{cccc}
        \toprule
        Resolution & Models & 1D Sinusoid & LOOPE (Ours)\\ 
        \midrule
        \multirow{3}{*}{\(224 \times 224\)} & ViT-Base \cite{vit} & 85.3\% & 88.1\% (+2.8\%) \\
        & ViT-Small \cite{vit}  &   81.6\% & 83.8\% ( +2.2\%)   \\
        & DeiT-Base \cite{deit} &    86.3\% & 89.8\% (\textbf{+3.5\%}) \\
        \midrule
        \multirow{3}{*}{\(384 \times 384\)} & ViT-Base \cite{vit} & 89.1\% & 92.2\% (+3.1\%)    \\
        & ViT-Small \cite{vit} &  83.0\% & 86.1\% (+3.1\%)   \\
        & DeiT-Base \cite{deit} &  88.5\% & 92.4\% (\textbf{+3.9\%})   \\
        \bottomrule
    \end{tabular}
    }
    \caption{Accuracy Comparison of different ViT models with Sinusoid and LOOPE PEs for different Image Resolutions on Oxford-IIIT. For both cases, patch shape is \(16\times16\).}
    \label{tab:resolution}
\end{table}
Table \ref{tab:resolution} presents the performance of different ViT models with Sinusoid and LOOPE PEs across two different resolutions. Our method shows higher improvement in accuracy with bigger resolution. 


\subsubsection{Visualization of Positional Encodings}
\begin{figure}[ht]
    \centering
    \begin{tabular}{ccc}
        \includegraphics[width=0.11\textwidth]{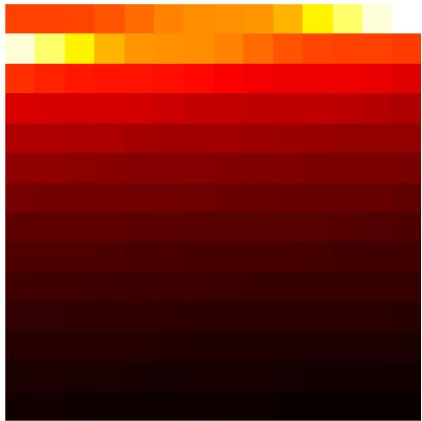} & 
        \includegraphics[width=0.11\textwidth]{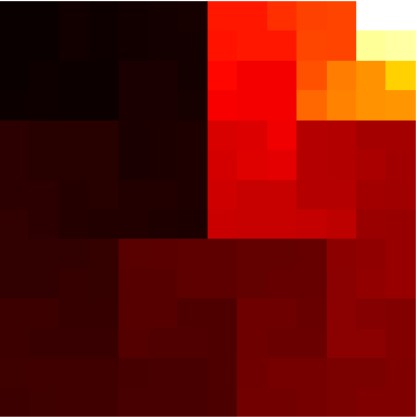} & 
        \includegraphics[width=0.11\textwidth]{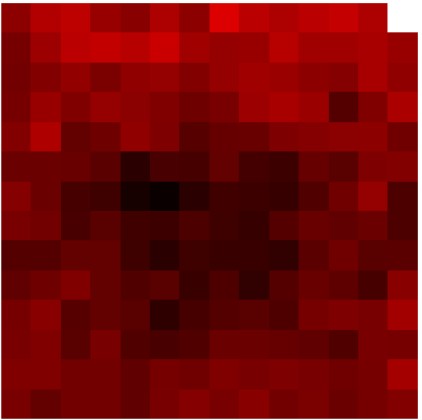} \\
        \includegraphics[width=0.11\textwidth]{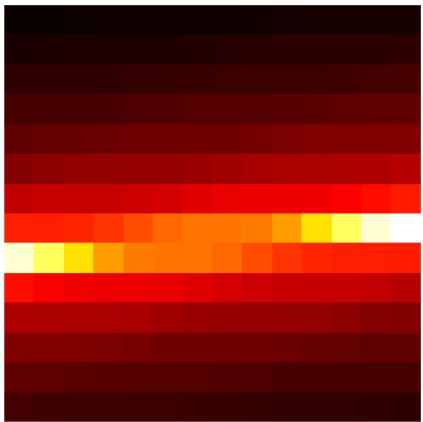} & 
        \includegraphics[width=0.11\textwidth]{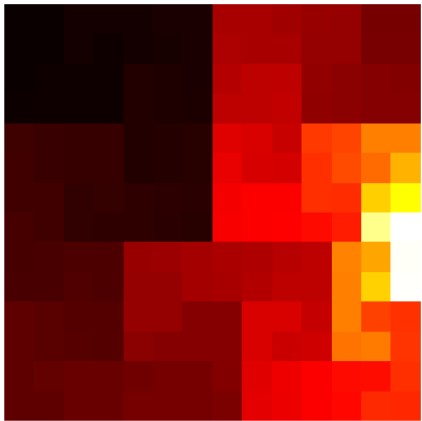} & 
        \includegraphics[width=0.11\textwidth]{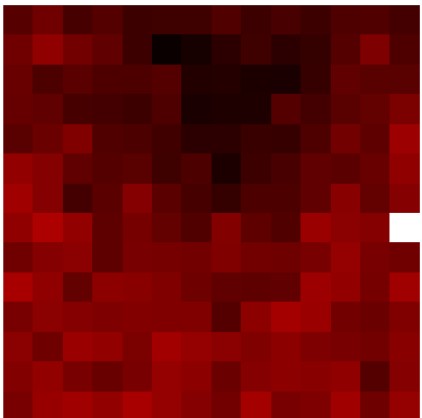} \\
        \includegraphics[width=0.11\textwidth]{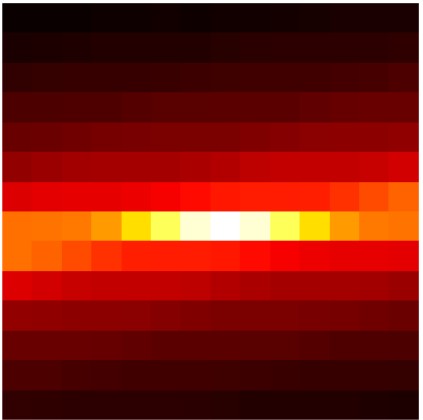} & 
        \includegraphics[width=0.11\textwidth]{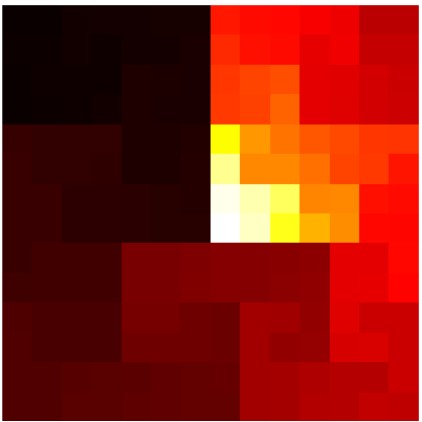} & 
        \includegraphics[width=0.11\textwidth]{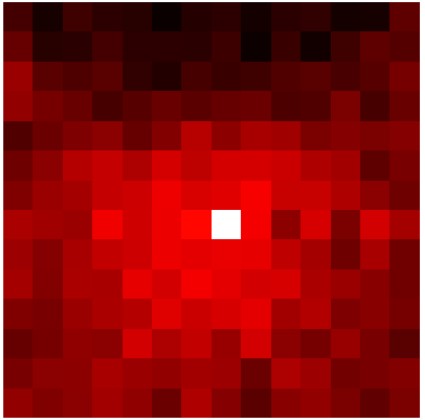} \\
        Sin APE & LOOPE & 1D Learnable \\
    \end{tabular}
    \caption{Cosine Similarity Maps for Three APEs: Top-Right Corner, Right-Boundary, and Middle Cell (Top to Bottom)}
    \label{fig:cosine}
\end{figure}

Figure \ref{fig:cosine} visualizes $14\times14$ cosine similarity maps for corner (Row 1), boundary (Row 2), and center (Row 3) reference cells. Ideal PEs exhibit smooth, monotonically decaying concentric patterns radiating outward. Sinusoidal APE (left) shows irregular jumps due to zigzag scan reversals at row boundaries. Conversely, LOOPE (center) yields notably smoother, concentric patterns across all positions, confirming our context-aware Gilbert ordering faithfully preserves spatial neighborhoods. 1D learnable APE (right) exhibits the weakest structure with scattered anomalies, showing that embeddings lacking geometric priors fail to maintain consistent spatial relationships (additional visualizations in Supp. B.5).

\section{Conclusion}
We introduced \textbf{LOOPE}, a learnable framework for patch ordering that unifies geometric structure and context-aware adaptation, establishing reordering as a key design dimension with preserved geometry and frequency robustness. In addition, we proposed the \textbf{Three-Cell experiment} and \textbf{PESI metrics} as principled tools for evaluating PEs beyond downstream accuracy. This perspective opens new directions for rethinking positional information in transformers. 

{
    \small
    \bibliographystyle{ieeenat_fullname}
    \bibliography{main}
}





\raggedbottom

\renewcommand{\topfraction}{0.9}
\renewcommand{\bottomfraction}{0.9}
\renewcommand{\textfraction}{0.1}
\renewcommand{\floatpagefraction}{0.8}
\setcounter{topnumber}{3}
\setcounter{bottomnumber}{3}
\setcounter{totalnumber}{5}

\title{LOOPE: Learnable Optimal Patch Order for Positional Encoders in Vision Transformers (Supplementary Material)}
\author{Anonymous CVPR: 2026 submission}
\date{}

\maketitlesupplementary

\renewcommand{\thesection}{\Alph{section}}
\section{Methodology}

\subsection{Gilbert Curve Generation (Generalized Hilbert Curve)}

The following pseudocode represents a recursive algorithm for generating a generalized Hilbert (Gilbert) space-filling curve for arbitrary 2D rectangular grids. The algorithm outputs discrete 2D coordinates that fill a rectangle of given width and height.

\subsubsection{Main Procedure: \texttt{gilbert2d}}

\begin{algorithm}[!t]
\caption{\texttt{gilbert2d(width, height)}}
\begin{algorithmic}[1]
\Function{gilbert2d}{$width,\; height$}
    \If{$width \ge height$}
        \State \Return \Call{generate2d}{$0,\; 0,\; width,\; 0,\; 0,\; height$}
    \Else
        \State \Return \Call{generate2d}{$0,\; 0,\; 0,\; height,\; width,\; 0$}
    \EndIf
\EndFunction
\end{algorithmic}
\end{algorithm}

\subsubsection{Helper Function: \texttt{sgn}}
\begin{algorithm}[htbp]
\caption{\texttt{sgn(x)}}
\begin{algorithmic}[1]
\Function{sgn}{$x$}
    \If{$x < 0$}
        \State \Return $-1$
    \ElsIf{$x > 0$}
        \State \Return $1$
    \Else
        \State \Return $0$
    \EndIf
\EndFunction
\end{algorithmic}
\end{algorithm}

\subsubsection{Recursive Procedure: \texttt{generate2d}}

\begin{algorithm}[!t]
\caption{\texttt{generate2d(x, y, ax, ay, bx, by)}}
\begin{algorithmic}[1]
\Procedure{generate2d}{$x, y, ax, ay, bx, by$}
    \State $w \gets |ax + ay|$ \Comment{Effective width}
    \State $h \gets |bx + by|$ \Comment{Effective height}
    \State $(dax,\; day) \gets (\text{sgn}(ax),\; \text{sgn}(ay))$ \Comment{Unit vector in major direction}
    \State $(dbx,\; dby) \gets (\text{sgn}(bx),\; \text{sgn}(by))$ \Comment{Unit vector in orthogonal direction}
    
    \If{$h = 1$}
        \For{$i \gets 0$ \textbf{to} $w-1$}
            \State \textbf{Output} $(x, y)$
            \State $(x, y) \gets (x + dax,\; y + day)$
        \EndFor
        \State \Return
    \EndIf
    
    \If{$w = 1$}
        \For{$i \gets 0$ \textbf{to} $h-1$}
            \State \textbf{Output} $(x, y)$
            \State $(x, y) \gets (x + dbx,\; y + dby)$
        \EndFor
        \State \Return
    \EndIf
    
    \State $(ax2,\, ay2) \gets \left(\lfloor ax/2 \rfloor,\; \lfloor ay/2 \rfloor\right)$
    \State $(bx2,\, by2) \gets \left(\lfloor bx/2 \rfloor,\; \lfloor by/2 \rfloor\right)$
    \State $w2 \gets |ax2 + ay2|$
    \State $h2 \gets |bx2 + by2|$
    
    \If{$2w > 3h$} \Comment{Long case: split into two parts}
        \If{$(w2 \bmod 2 = 1)$ \textbf{and} $(w > 2)$}
            \State $(ax2,\, ay2) \gets (ax2 + dax,\; ay2 + day)$ \Comment{Prefer even steps}
        \EndIf
        \State \Call{generate2d}{$x,\; y,\; ax2,\; ay2,\; bx,\; by$}
        \State \Call{generate2d}{$x + ax2,\; y + ay2,\; ax - ax2,\; ay - ay2,\; bx,\; by$}
    \Else
        \If{$(h2 \bmod 2 = 1)$ \textbf{and} $(h > 2)$}
            \State $(bx2,\, by2) \gets (bx2 + dbx,\; by2 + dby)$ \Comment{Prefer even steps}
        \EndIf
        \State \Call{generate2d}{$x,\; y,\; bx2,\; by2,\; ax2,\; ay2$}
        \State \Call{generate2d}{$x + bx2,\; y + by2,\; ax,\; ay,\; bx - bx2,\; by - by2$}
        \State \Call{generate2d}{$x + (ax-dax) + (bx2-dbx),\, y + (ay-day) + (by2-dby),\; -bx2,\; -by2,\; -(ax-ax2),\; -(ay-ay2)$}
    \EndIf
\EndProcedure
\end{algorithmic}
\end{algorithm}

\textbf{Notes:}
\begin{itemize}
    \item The function \texttt{gilbert2d} selects the major direction based on the aspect ratio.
    \item The helper function \texttt{sgn} returns the sign of its input.
    \item The \texttt{generate2d} procedure is recursive. It subdivides the rectangle until reaching trivial row or column fills, outputting coordinate points at each base case.
    \item In this pseudocode, ``\textbf{Output} $(x, y)$'' corresponds to yielding the coordinate in the Python implementation.
\end{itemize}

\subsection{Context-Aware Index Generator Architecture}
\begin{table*}[t]
\centering
\small
\caption{Layer-wise specification of the context-aware index generator for $X_C$. Conv$_k$ denotes a convolution with kernel size $k \times k$. All convolutions use ReLU activations unless stated otherwise.}
\label{tab:xc-arch}
\begin{tabular}{|c|c|p{3.8cm}|c|p{3.5cm}|}
\hline
\textbf{Layer} & \textbf{Input Shape} & \textbf{Operation} & \textbf{Output Shape} & \textbf{Notes} \\
\hline
Input & $[5 \times H \times W]$ & \textemdash & $[5 \times H \times W]$ & RGB + coordinate maps \\
\hline
Conv1 & $[5 \times H \times W]$ & 32 filters, kernel $P \times P$, stride $16$ & $[32 \times h \times w]$ & Downsampling to patch scale \\
\hline
Conv2 & $[32 \times h \times w]$ & 16 filters, kernel $5 \times 5$, stride $1$ & $[16 \times h \times w]$ & Local refinement \\
\hline
Conv3 & $[16 \times h \times w]$ & 8 filters, kernel $5 \times 5$, stride $1$ & $[8 \times h \times w]$ & \textemdash \\
\hline
Conv4 & $[8 \times h \times w]$ & 4 filters, kernel $5 \times 5$, stride $1$ & $[4 \times h \times w]$ & \textemdash \\
\hline
Conv5 & $[4 \times h \times w]$ & 1 filter, kernel $5 \times 5$, stride $1$ & $[1 \times h \times w]$ & Channel squeeze \\
\hline
BN + Flatten & $[1 \times h \times w]$ & BatchNorm + reshape & $[1 \times N]$ & $N = h \times w$ patches \\
\hline
MLP & $[1 \times N]$ & Linear layer + bias & $[1 \times N]$ & Fully connected refinement \\
\hline
Activation & $[1 \times N]$ & $2\sigma(x) - 1$ & $X_C \in [-1,1]^{1 \times N}$ & Fractional offsets \\
\hline
\end{tabular}
\end{table*}
The architecture of the context-aware bias generator $X_C$ is summarized in Table~\ref{tab:xc-arch}. The module takes as input the RGB image (\(I_0 \in \mathbf{R}^{3 \times H \times W} \)) concatenated with coordinate maps ($x,y \in \mathbf{R}^{1 \times H \times W}$), forming a $5$-channel tensor. Through a series of convolutional layers followed by batch normalization, flattening, and an MLP, the network outputs continuous offsets $X_C \in \mathbf{R}^{1 \times N}$, which refine the static patch order $X_G$.

The stride of 16 in Conv1 ensures that each feature corresponds to a patch region of the image, aligning $X_C$ directly with the patch grid used by the transformer. Subsequent layers refine this representation without further spatial downsampling, keeping the module lightweight and efficient.

\subsection{Theoretical Effect of Context-Aware Adaptation on Frequency Sensitivity}

To analyze why the context-aware refinement \(X_C\) reduces sensitivity to the choice of frequency set, consider a simplified form of the positional encoding:  
\begin{equation}
    \mathrm{PE}(f) = \sin\!\left( \frac{(x + \delta)}{N} \, f \right),
    \label{eq:pe_simplified}
\end{equation}
where \(x\) is the static patch index, \(\delta\) is the adaptive offset produced by the context-aware generator, \(N\) is the normalization factor (e.g., \(N=196\) for a \(14 \times 14\) patch grid), and \(f\) is the frequency. The offset \(\delta\) itself depends on both the frequency and local image context:  
\[
\delta = g(f, \text{context}).
\]

\paragraph{Frequency sensitivity without adaptation.}  
When \(\delta = 0\), the derivative of \(\mathrm{PE}\) with respect to \(f\) is  
\begin{equation}
    \frac{\partial}{\partial f}\,\mathrm{PE}(f) 
    = \frac{x}{N} \cos\!\left( \frac{x}{N} f \right).
    \label{eq:deriv_noadapt}
\end{equation}
Thus, sensitivity is strictly proportional to the normalized index \(x/N\). In this case, different choices of frequency sets directly translate into large oscillations in the embedding, making performance highly dependent on design choices.

\paragraph{Frequency sensitivity with adaptation.}  
With context-aware refinement (\(\delta \neq 0\)), we compute
\begin{align}
    \frac{\partial}{\partial f}\,\mathrm{PE}(f) 
    &= \frac{\partial}{\partial f}\,\sin\!\left( \frac{(x+\delta)}{N} f \right) \nonumber \\
    &= \cos\!\left( \frac{(x+\delta)}{N} f \right) \cdot 
       \frac{\partial}{\partial f}\!\left( \frac{(x+\delta)}{N} f \right). 
    \label{eq:deriv_expand}
\end{align}

Expanding the inner derivative:
\begin{equation}
    \frac{\partial}{\partial f}\!\left( \frac{(x+\delta)}{N} f \right) 
    = \frac{x+\delta}{N} + \frac{f}{N}\,\frac{\partial \delta}{\partial f}.
    \label{eq:inner_deriv}
\end{equation}

Substituting back into Eq.~\ref{eq:deriv_expand} gives the simplified expression:
\begin{equation}
    \frac{\partial}{\partial f}\,\mathrm{PE}(f) 
    = \left( \frac{x+\delta}{N} + \frac{f}{N}\,\frac{\partial \delta}{\partial f} \right)
      \cos\!\left( \frac{(x+\delta)}{N} f \right).
    \label{eq:deriv_simplified}
\end{equation}

\paragraph{Interpretation.}  
Equation~\ref{eq:deriv_simplified} reveals two contributions to frequency sensitivity:
\begin{itemize}
    \item \(\tfrac{x+\delta}{N}\): a shifted static term, analogous to the non-adaptive case but now adjusted by the learned offset \(\delta\).  
    \item \(\tfrac{f}{N}\,\tfrac{\partial \delta}{\partial f}\): an adaptive correction term, which enables the model to compensate for changes in the frequency set.  
\end{itemize}
Without adaptation, only the first term exists, making sensitivity rigidly determined by \(x/N\). With adaptation, the second term actively counterbalances frequency variations, thereby reducing sensitivity while preserving the sinusoidal form of the encoding.

\paragraph{Conclusion.}  
Context-aware adaptation therefore moderates the dependency on frequency selection: the encoding remains sinusoidal in structure, ensuring consistency, but the added corrective term \(\tfrac{\partial \delta}{\partial f}\) provides robustness. This theoretical view aligns with the empirical results in Table. 7, where LOOPE maintains stable accuracy even under non-optimal frequency choices.

\subsection{Three Cell experiment}
\subsubsection*{Overview}
This dataset is designed to facilitate research in spatial reasoning, geometric transformations, and pattern recognition using synthetic images. It consists of structured grid-based images where three distinct color markers (red, green, and blue) are positioned according to predefined spatial constraints.

\subsubsection*{Purpose \& Applications}
The dataset supports tasks such as:
\begin{itemize}
    \item \textbf{Machine Learning \& Deep Learning:} Training models to understand spatial relationships.
    \item \textbf{Computer Vision:} Evaluating geometric transformations and positional reasoning.
    \item \textbf{Representation Learning:} Analyzing how models interpret structured spatial layouts.
\end{itemize}

\textbf{Dataset Statistics and Description}

\begin{itemize}
    \item The dataset consists of 10,000 synthetic images.
    \item The cell coordinates are uniformly sampled on a $14 \times 14$ grid.
    \item The relative cases are designed to be equiprobable, ensuring balanced distribution.
    \item Among the 196 patches, 193 are colored black; the three colored patches are non-colinear and non-overlapping, however they can share boundaries.
\end{itemize}

\textbf{Three-Cell Experiment Dataset Visualization}

\begin{figure}[H]
  \centering
  \begin{subfigure}{0.2\linewidth}
    \includegraphics[width=\linewidth]{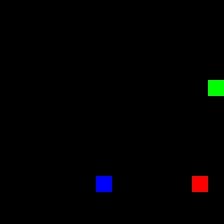}
    \caption{}
    \label{fig: distance}
  \end{subfigure}
  \begin{subfigure}{0.2\linewidth}
    \includegraphics[width=\linewidth]{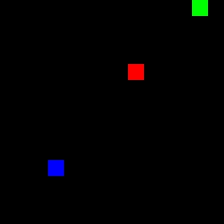}
    \caption{}
    \label{fig:orientation}
  \end{subfigure}
  \begin{subfigure}{0.2\linewidth}
    \includegraphics[width=\linewidth]{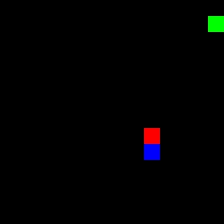}
    \caption{}
    \label{fig:area}
  \end{subfigure}
  \begin{subfigure}{0.2\linewidth}
    \includegraphics[width=\linewidth]{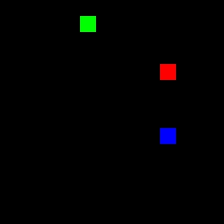}
    \caption{}
    \label{fig:vector}
  \end{subfigure}  
  \caption{Three-Cell Experiment Sample Dataset}
  \label{fig:three-pixel-experiment0}
\end{figure}

\textbf{Structure \& Content}
Each sample includes:
\begin{enumerate}
    \item \textbf{Image:} A $224 \times 224$ pixel RGB image that represents a $14 \times 14$ grid. In this grid:
    \begin{itemize}
        \item A red cell serves as the reference point.
        \item A green cell and a blue cell are positioned based on mathematical constraints.
        \item The background is black, with each cell having a distinct color.
    \end{itemize}
    \item \textbf{Binary Mask:} A 6-element vector encoding specific spatial properties, such as:
    \begin{itemize}
        \item Distance comparisons.
        \item Geometric orientation.
        \item Area relationships.
    \end{itemize}
\end{enumerate}
\textbf{Key Characteristics}
\begin{itemize}
    \item \textbf{Controlled Complexity:} Images are generated based on strict mathematical rules.
    \item \textbf{Diversity:} Multiple spatial configurations ensure a wide range of positional relationships.
    \item \textbf{Explainability:} The structured nature of the dataset makes it ideal for interpretable AI research.
\end{itemize}
\textbf{Potential Use Cases}
\begin{itemize}
    \item Training AI models for spatial awareness and geometric reasoning.
    \item Benchmarking representation learning techniques in structured visual tasks.
    \item Investigating how neural networks learn spatial relationships in a controlled setting.
\end{itemize}

\begin{algorithm}[H]
\scriptsize
\caption{Synthetic Image Dataset Generation}
\begin{algorithmic}[1]
\Procedure{GenerateSyntheticDataset}{}
    \State \textbf{Input:} Grid size $G \gets 14$, Sum-of-squares limit $L \gets 338$, $max\_attempts$
    \State \textbf{Output:} Expanded grid image $grid_{224}$ and binary mask $mask[6]$
    \State $max\_val \gets \lfloor \sqrt{L} \rfloor$
    \State Initialize dictionary $D \gets \emptyset$
    \For{$a = -max\_val$ \textbf{to} $max\_val$}
        \For{$b = -max\_val$ \textbf{to} $max\_val$}
            \State $n \gets a^2 + b^2$
            \If{$n \leq L$}
                \State Append pair $(a, b)$ to $D[n]$
            \EndIf
        \EndFor
    \EndFor
    \State Remove keys from $D$ that have only one representation
    \State Randomly choose mode from $\{1,2,3\}$
    \State $valid \gets \textbf{false}$
    \While{not $valid$}
        \State Randomly select red cell $X_r \in [0,G-1]^2$
        \If{mode $= 1$}
            \State Select random key $k$ from keys of $D$
            \For{$attempt = 1$ to $max\_attempts$}
                \State Randomly choose two distinct pairs $(a_1,b_1)$ and $(a_2,b_2)$ from $D[k]$
                \State $X_g \gets X_r + (a_1,b_1)$
                \State $X_b \gets X_r + (a_2,b_2)$
                \If{$X_g$ and $X_b$ are within the grid and distinct from $X_r$ and each other}
                    \State $valid \gets \textbf{true}$
                    \State \textbf{break}
                \EndIf
            \EndFor
        \ElsIf{mode $\in \{2,3\}$}
            \For{$attempt = 1$ to $max\_attempts$}
                \State Randomly select two keys $k_1,k_2$ from $D$
                \If{mode $= 2$}
                    \State $k_g \gets \min(k_1, k_2)$, $k_b \gets \max(k_1, k_2)$
                \Else
                    \State $k_g \gets \max(k_1, k_2)$, $k_b \gets \min(k_1, k_2)$
                \EndIf
                \State Randomly select pair $(a_g,b_g)$ from $D[k_g]$ and $(a_b,b_b)$ from $D[k_b]$
                \State $X_g \gets X_r + (a_g,b_g)$
                \State $X_b \gets X_r + (a_b,b_b)$
                \If{$X_g$ and $X_b$ are within the grid and distinct}
                    \State $valid \gets \textbf{true}$
                    \State \textbf{break}
                \EndIf
            \EndFor
        \EndIf
    \EndWhile
    \State Create a $G \times G$ grid $grid_{14}$ with black background
    \State Set $grid_{14}[X_r] \gets$ red, $grid_{14}[X_g] \gets$ green, $grid_{14}[X_b] \gets$ blue
    \State Expand $grid_{14}$ to $grid_{224}$ by replicating each cell into a $16 \times 16$ block
    \State Initialize mask $mask[6] \gets (0,0,0,0,0,0)$
    \State \textbf{Call:} \textsc{DistanceCompare}, \textsc{Orientation}, \textsc{AreaCompare}, \textsc{VectorSum} on $(X_r, X_g, X_b, mask)$
    \State \textbf{return} $grid_{224}$ and $mask$
\EndProcedure
\end{algorithmic}
\end{algorithm}

\subsection{Positional Embeddings Structural Integrity (PESI) Metrics}

\subsubsection{Undirected Monotonicity}

\begin{algorithm}[H]
\footnotesize
\caption{Compute Global Undirected Monotonicity Score \( M_u \)}
\textbf{Input:} Positional embedding tensor \( P \in \mathbb{R}^{H \times W \times D} \)\\
\textbf{Output:} Global undirected monotonicity score \( M_u \)
\begin{algorithmic}[1]
\State Initialize an empty list \( S \gets [\,] \)
\For{\( x = 1,2,\ldots,H \)}
    \For{\( y = 1,2,\ldots,W \)}
        \State \( v_{\text{center}} \gets P(x,y,:) \)
        \State Compute norm: \( \|v_{\text{center}}\| \gets \text{norm}(v_{\text{center}}) + \epsilon \)
        \State Compute cosine similarity for all \( (i,j) \):
        \( A_{(x,y)}(i,j) \gets \frac{v_{\text{center}} \cdot P(i,j,:)}{\|v_{\text{center}}\| \cdot \|P(i,j,:)\| + \epsilon} \)
        \State Compute Euclidean distances:
        \( d(i,j) \gets \sqrt{(i-x)^2 + (j-y)^2} \)
        \State Form radial bins: \( r(i,j) \gets \text{round}\bigl(d(i,j)\bigr) \)
        \For{each unique radial bin \( r \)}
            \State \( S_{(x,y)}(r) \gets \) average of \( A_{(x,y)}(i,j) \) for all \( (i,j) \) with \( r(i,j)=r \)
        \EndFor
        \State Compute Spearman's rank correlation \( \rho_{(x,y)} \) between \( \{r\} \) and \( \{S_{(x,y)}(r)\} \)
        \State Append \( \rho_{(x,y)} \) to \( S \)
    \EndFor
\EndFor
\State Compute global score:
\( M_u \gets \frac{1}{HW} \sum_{(x,y)} \bigl( 1 - \rho_{(x,y)} \bigr) \)
\State \Return \( M_u \)
\end{algorithmic}
\end{algorithm}

\subsubsection{Directed Monotonicity}

\begin{algorithm}[H]
\footnotesize
\caption{Compute Global Directed Monotonicity Measure \(M_D\)}
\textbf{Input:} Positional embedding tensor \(P \in \mathbb{R}^{H \times W \times D}\), quantization angle \(\delta\)\\
\textbf{Output:} Global directed monotonicity measure \(M_D\)
\begin{algorithmic}[1]
\State \(N \gets \lceil 2\pi/\delta \rceil\) \Comment{Total number of directional buckets}
\State Initialize list \(S \gets [\,]\)
\For{\(x = 1,2,\ldots,H\)}
    \For{\(y = 1,2,\ldots,W\)}
        \State \(v_{\text{center}} \gets P(x,y,:)\)
        \State Cosine similarity: \(A_{(x,y)}(i,j) \gets \frac{v_{\text{center}} \cdot P(i,j,:)}{\|v_{\text{center}}\|\,\|P(i,j,:)\| + \epsilon}\)
        \State Radial distances: \(d(i,j) \gets \sqrt{(i-x)^2 + (j-y)^2}\)
        \State Angles: \(\theta_{(x,y)}(i,j) \gets \operatorname{atan2}(j-y,\,i-x)\)
        \State Quantize: \(k(i,j) \gets \lfloor \theta_{(x,y)}(i,j)/\delta \rfloor \bmod N\)
        \State Initialize list \(\rho^k \gets [\,]\)
        \For{\(k = 0,1,\ldots,N-1\)}
            \State Let \(B_k \gets \{(i,j) \mid k(i,j)=k\}\)
            \If{\(|B_k| < 2\)}
                \State Append \(0\) to \(\rho^k\)
            \Else
                \State Order \(B_k\) by increasing \(d(i,j)\)
                \State Compute similarity profile \(S_{(x,y)}^k(r)\) for ordered cells
                \State Spearman: \(\rho_{(x,y)}^k \gets 1 - \frac{6\sum_{r} d_r^2}{|B_k|(|B_k|^2-1)}\)
                \State Append \(\rho_{(x,y)}^k\) to \(\rho^k\)
            \EndIf
        \EndFor
        \State Mean correlation: \(\bar{\rho}_{(x,y)} \gets \frac{1}{N}\sum_{k=0}^{N-1}\rho^k\)
        \State Append \(1 - \bar{\rho}_{(x,y)}\) to \(S\)
    \EndFor
\EndFor
\State Global measure: \(M_D \gets \frac{1}{HW}\sum_{(x,y)} \bigl(1-\bar{\rho}_{(x,y)}\bigr)\)
\State \Return \(M_D\)
\end{algorithmic}
\end{algorithm}

\subsubsection{Undirected Asymmetry}

\begin{algorithm}[H]
\footnotesize
\caption{Compute Global Undirected Asymmetry \(A_{SU}\)}
\textbf{Input:} Positional embedding tensor \(P \in \mathbb{R}^{H \times W \times D}\)\\
\textbf{Output:} Global undirected asymmetry measure \(A_{SU}\)
\begin{algorithmic}[1]
\For{each center \((x,y)\) in \(\{1,\dots,H\} \times \{1,\dots,W\}\)}
    \State Cosine similarity: \(A_{(x,y)}(i,j) = \frac{P(x,y,:) \cdot P(i,j,:)}{\|P(x,y,:)\| \, \|P(i,j,:)\|}\)
    \State Euclidean distance: \(d(i,j) = \sqrt{(i-x)^2 + (j-y)^2}\)
    \State Radial bins: \(B_r = \{ (i,j) \mid r = \text{round}(d(i,j)) \}\)
    \For{each radial bin \(r \in R\)}
        \State Mean similarity: \(\mu_{(x,y)}(r) = \frac{1}{|B_r|} \sum_{(i,j) \in B_r} A_{(x,y)}(i,j)\)
        \State Std.\ deviation: \(\sigma_{(x,y)}(r) = \sqrt{ \frac{1}{|B_r|} \sum_{(i,j) \in B_r} \bigl( A_{(x,y)}(i,j) - \mu_{(x,y)}(r) \bigr)^2 }\)
        \State Coeff.\ of variation: \(\mathrm{CV}_{(x,y)}(r) = \sigma_{(x,y)}(r) / \mu_{(x,y)}(r)\)
    \EndFor
    \State Undirected asymmetry: \(A'_{SU}(x,y) = \frac{1}{|R|} \sum_{r \in R} \mathrm{CV}_{(x,y)}(r)\)
\EndFor
\State Global measure: \(A_{SU} = \frac{1}{HW} \sum_{x=1}^{H} \sum_{y=1}^{W} A'_{SU}(x,y)\)
\State \Return \(A_{SU}\)
\end{algorithmic}
\end{algorithm}

\subsection{Sensitivity Calculation Documentation}

We consider four input signals of length \(L = 768\):
\begin{equation}
    f_1(i) = 0.978^i,
\end{equation}
\begin{equation}
    f_2(i) = 1 - i \frac{1 - 0.0001}{L - 1},
\end{equation}
\begin{equation}
    f_3(i) = 0.9^i,
\end{equation}
\begin{equation}
    f_4(i) \sim \text{Uniform}(0.0001, 1) , \text{sorted descending}.
\end{equation}

The first signal \(f_1\) is chosen as the baseline reference.

\subsubsection{Relative Change of Input Signals}
For each signal \(f_j\) (\(j=2,3,4\)), its relative change compared to the baseline \(f_1\) is computed as the normalized root mean squared difference:
\begin{equation}
    \Delta f_j = \frac{\lVert f_j - f_1 \rVert_2}{\lVert f_1 \rVert_2},
\end{equation}
where \( \lVert \cdot \rVert_2 \) denotes the Euclidean norm.

\subsubsection{Output Values}
Each black-box algorithm produces four output values corresponding to the four input signals. Denote the outputs for algorithm \(k\) as
\begin{equation}
    O^{(k)} = [O^{(k)}_1, O^{(k)}_2, O^{(k)}_3, O^{(k)}_4],
\end{equation}
where \(O^{(k)}_1\) is the output corresponding to the baseline \(f_1\).

\subsubsection{Sensitivity per Input}
For each algorithm \(k\) and each signal \(j=2,3,4\), the sensitivity is defined as the ratio of the absolute output change to the relative input change:
\begin{equation}
    S^{(k)}_j = \frac{|O^{(k)}_j - O^{(k)}_1|}{\Delta f_j}.
\end{equation}

\subsubsection{Normalized Average Sensitivity}
Since the relative changes \(\Delta f_j\) may differ significantly across signals, we compute a normalized weighted average sensitivity for each algorithm:
\begin{equation}
    \bar{S}^{(k)} = 
    \frac{\sum_{j=2}^4 S^{(k)}_j  \Delta f_j}{\sum_{j=2}^4 \Delta f_j}.
\end{equation}

This measure reflects the overall sensitivity of the algorithm to input variations, properly weighted by the magnitude of the input changes.

\section{Additional Experiments and Visualization}

\setcounter{table}{1}

\subsection{Training from Scratch on ImageNet-100}

A natural question is whether LOOPE's gains are tied to pretrained weight initialization or reflect a genuine structural advantage of the positional encoding. To answer this, we train ViT-Base and Cross-ViT entirely from scratch on ImageNet-100 using an identical training protocol (learning rate schedule, augmentation, optimizer) across all positional encoding baselines. No pretrained weights are used for any method. Table~\ref{tab:scratch} reports the best top-1 accuracy achieved and the epoch at which it occurred. The results confirm that the performance ranking observed in the main paper (Table~1) is preserved under from-scratch training: LOOPE consistently achieves the highest accuracy on both architectures while converging within a comparable number of epochs to the other methods. This demonstrates that the benefit of LOOPE stems from its encoding structure rather than any interaction with pretrained representations.

\begin{table}[H]
\centering
\small
\setlength{\tabcolsep}{3.5pt}
\begin{tabular}{lccccc}
\toprule
Model & No PE & Learnable & 1D Sin & \makecell{Static} & \makecell{LOOPE} \\
\midrule
\makecell{ViT-Base \\ (Best Epochs)} & \makecell{87.1\% \\ 258} & \makecell{87.9\% \\ 262} & \makecell{88.8\% \\ 280} & \makecell{89.0\% \\ 265} & \makecell{\textbf{89.2\%} \\ 267} \\
\midrule
\makecell{Cross-ViT \\ (Best Epochs)} & \makecell{86.6\% \\ 268} & \makecell{87.7\% \\ 271} & \makecell{87.9\% \\ 293} & \makecell{88.8\% \\ 275} & \makecell{\textbf{89.4\%} \\ 277} \\
\bottomrule
\end{tabular}
\caption{From-scratch training on ImageNet-100 shows a similar trend as Table~1 in the main paper. LOOPE achieves the highest accuracy on both ViT-Base and Cross-ViT.}
\label{tab:scratch}
\end{table}

\subsection{3-Cell Input Ablation: RGB vs.\ Coordinate Channels}

The context-aware index generator $G$ receives both patch coordinates $(x,y)$ and RGB pixel values as input (Sec.~3.2 in the main paper). A key design question is whether the learned refinement $X_C$ relies on semantic (appearance) cues from the RGB channels or primarily on geometric (positional) information from the coordinate maps. To disentangle these contributions, we evaluate LOOPE on the 3-Cell benchmark under three input configurations: RGB channels only, coordinate maps only, and the full RGB+Coordinate input. Table~\ref{tab:3cell-ablation} reports the sub-task accuracies. The coordinate-only variant captures the majority of the gain over the static baseline across all four sub-tasks, while the RGB-only variant provides only a marginal improvement over the static ordering. The full RGB+Coordinate configuration yields the best results, indicating that the two signal types are complementary but that coordinate geometry is the dominant driver. This confirms that LOOPE's context-aware refinement is primarily geometry-driven, which is consistent with the theoretical motivation in Sec.~3.

\begin{table}[H]
\centering
\small
\setlength{\tabcolsep}{4pt}
\begin{tabular}{lcccc}
\toprule
PE 
& Dist. 
& Orient. 
& Area 
& Vec-Sum \\
\midrule
Static ($X_G$) 
& 88.8\% & 95.0\% & 89.9\% & 93.8\% \\
LOOPE (RGB only) 
& 89.9\% & 95.0\% & 91.6\% & 92.4\% \\
LOOPE (Coordinates only) 
& 91.9\% & 95.5\% & 92.8\% & 94.4\% \\
LOOPE (RGB+Coord.) 
& \textbf{93.4\%} & \textbf{95.8\%} & \textbf{93.3\%} 
& \textbf{94.6\%} \\
\bottomrule
\end{tabular}
\caption{3-Cell ablation: performance is largely explained by coordinate (geometry) input, not RGB semantics. Coordinate-only captures most of the gain over the Static baseline.}
\label{tab:3cell-ablation}
\end{table}

\subsection{PESI Metrics and 3-Cell Task Alignment}

The PESI metrics ($M_U$, $M_D$, $A_{SU}$) are intended as diagnostic probes of the similarity geometry induced by different positional encodings. To validate that these metrics capture meaningful structural properties, Table~\ref{tab:pesi-3cell} presents the PESI scores alongside the corresponding 3-Cell sub-task accuracies for each PE. The expected alignment is clearly visible: PEs with higher undirected monotonicity $M_U$ yield stronger distance reasoning accuracy, while higher directed monotonicity $M_D$ corresponds to stronger orientation accuracy. Notably, Learnable PE has the lowest scores on both $M_U$ and $M_D$, and correspondingly achieves the weakest distance and orientation accuracy in the 3-Cell benchmark. At the other end, LFF achieves the highest $M_D$ and orientation accuracy, while LOOPE achieves the highest $M_U$ and distance accuracy. The asymmetry metric $A_{SU}$ captures a different axis: the degree of directional bias in the similarity field. These results confirm that PESI metrics serve as compact, interpretable proxies for the geometric behaviors that matter in downstream positional reasoning.

\begin{table}[H]
\centering
\small
\setlength{\tabcolsep}{3.5pt}
\begin{tabular}{lccccc}
\toprule
PE 
& $M_U$ 
& Dist. 
& $M_D$ 
& Orient.
& $A_{SU}$ \\
\midrule
Learnable & \underline{1.7493} & \underline{85.6\%} & \underline{1.2003} & \underline{89.3\%} & $-$0.7272 \\
1D Sin & 1.9567 & 90.9\% & 1.4905 & 94.9\% & 0.1243 \\
LFF & 1.9623 & 93.1\% & \textbf{1.5230} & \textbf{96.7}\% & 0.2683 \\
Static $(X_G)$ & 1.9670 & 88.8\% & 1.2897 & 95.0\% & 0.0945 \\
\midrule
LOOPE $(X_G+X_c)$ & \textbf{1.9674} & \textbf{93.4\%} & 1.2900 & 95.8\% & 0.0939 \\
\bottomrule
\end{tabular}
\caption{Combined view of PESI metrics and 3-Cell sub-task accuracies. \underline{Underlined}: lowest; \textbf{Bold}: highest. Higher $M_U$ aligns with stronger distance accuracy; higher $M_D$ aligns with stronger orientation accuracy.}
\label{tab:pesi-3cell}
\end{table}

\subsection{Trends in PESI metrics}

\begin{figure}[ht]
    \centering
    \includegraphics[width=0.40\textwidth]{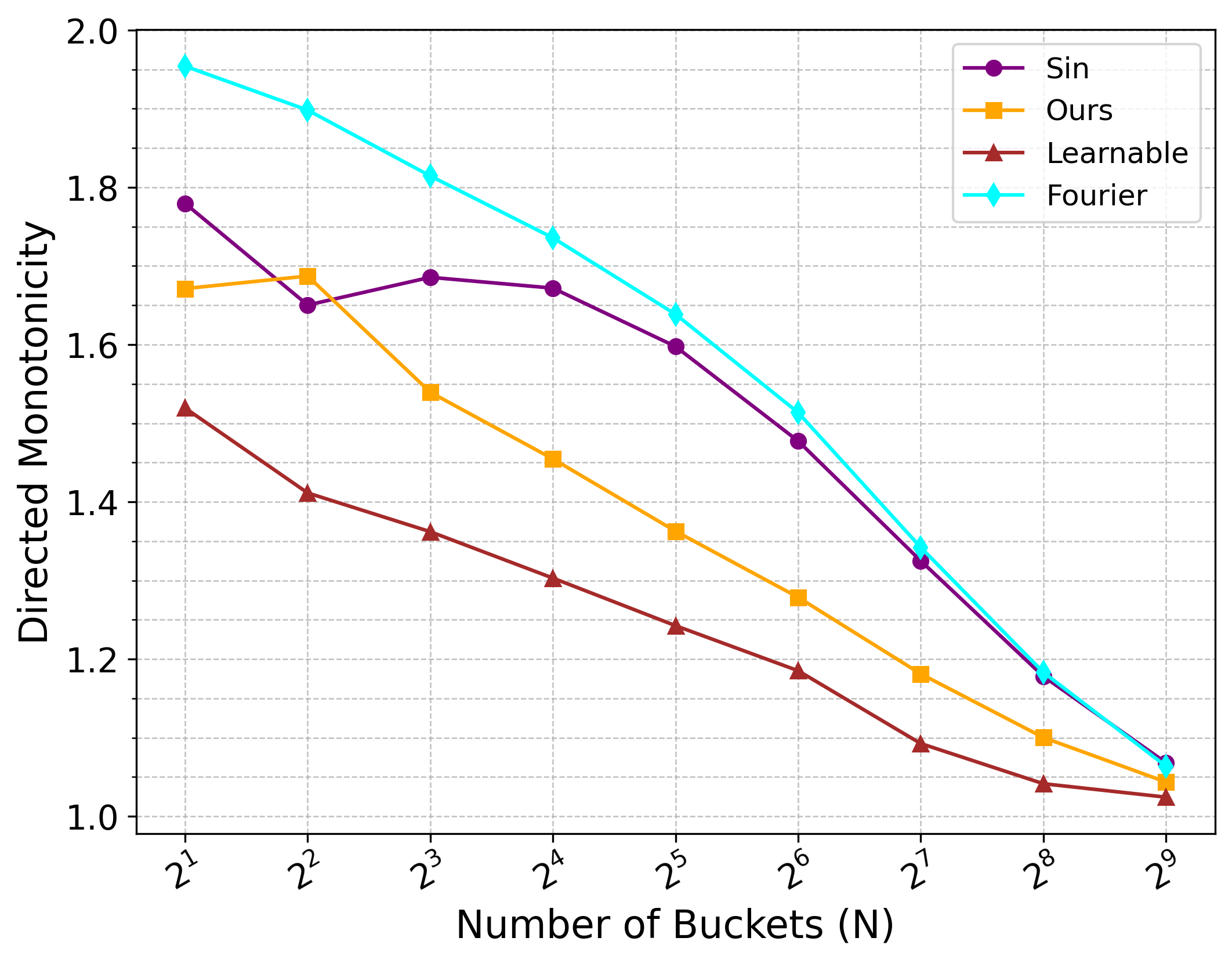}
    \caption{Trend of directed Monotonicity, \(M_D\) with increasing angle precision, \(\delta=2\pi/N\)}
    \label{fig:md}
\end{figure}
Figure \ref{fig:md}, shows the interesting facts about directed monotonicity, with this tool one can investigate how precisely the positional embeddings can maintain monotonicity. Clearly, increasing precision all positional encoders struggles to provide a monotone trend in cosine similarity. As \(N\rightarrow 1, M_D \rightarrow M_U\) which is exactly what we expected. At \(N=4,\delta=\pi/2,\) we can see that, LOOPE outperforms zigzag and learnable as it highly depends on hilbert order which propagates in square pattern.

\subsection{Visualization}

\begin{figure}[t]
    \centering
    \setlength{\tabcolsep}{1pt}
    \renewcommand{\thesubfigure}{}
    \begin{tabular}{ccccc} 
    &\textbf{Sin} & \textbf{Learnable} & \textbf{Fourier} & \textbf{Ours} \\  
    \textbf{(0,0)} & \subfloat{\includegraphics[width=0.1\textwidth]{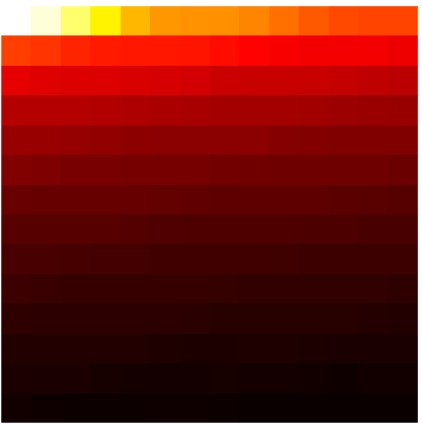}} &
    \subfloat{\includegraphics[width=0.1\textwidth]{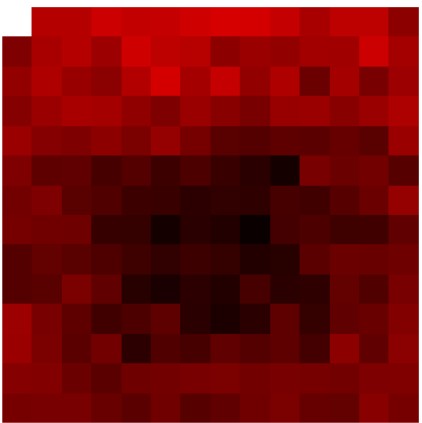}} &
    \subfloat{\includegraphics[width=0.1\textwidth]{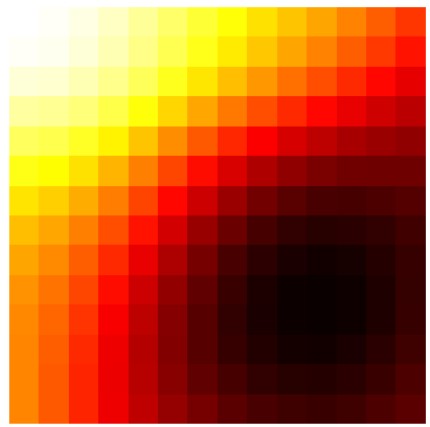}} &
    \subfloat{\includegraphics[width=0.1\textwidth]{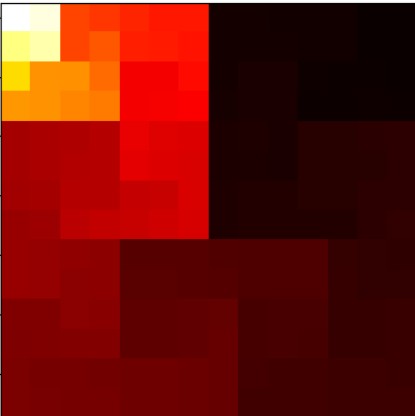}} \\

    \textbf{(13,0)} & \subfloat{\includegraphics[width=0.1\textwidth]{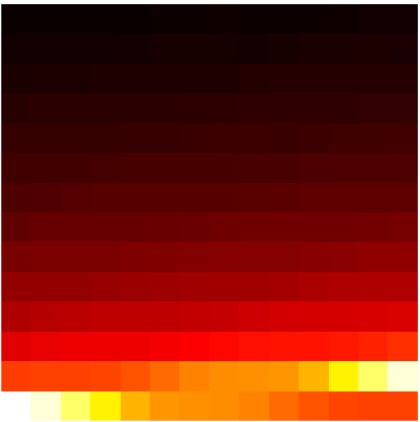}} &
    \subfloat{\includegraphics[width=0.1\textwidth]{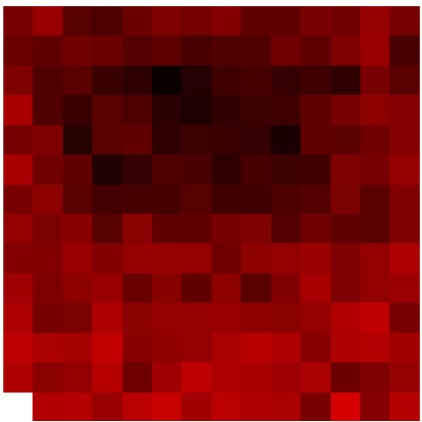}} &
    \subfloat{\includegraphics[width=0.1\textwidth]{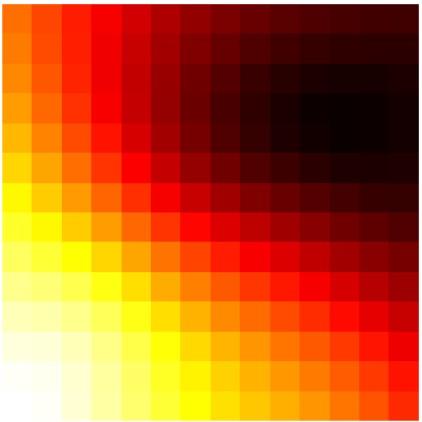}} &
    \subfloat{\includegraphics[width=0.1\textwidth]{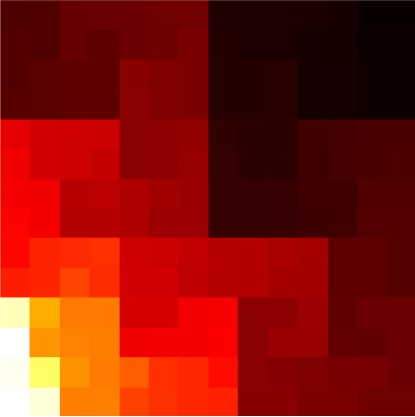}} \\

    \textbf{(0,13)} & \subfloat{\includegraphics[width=0.1\textwidth]{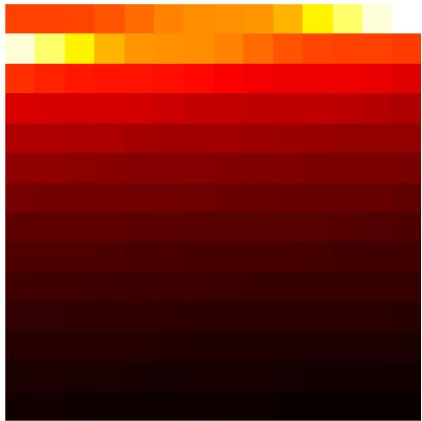}} &
    \subfloat{\includegraphics[width=0.1\textwidth]{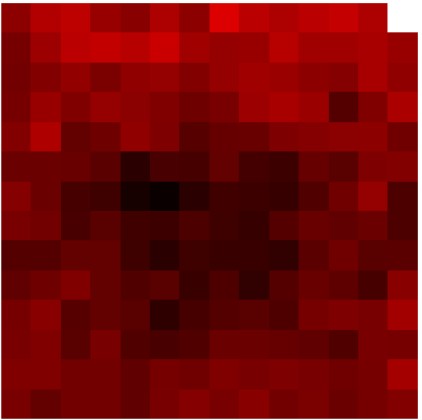}} &
    \subfloat{\includegraphics[width=0.1\textwidth]{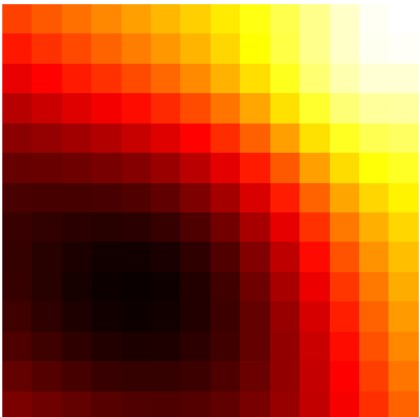}} &
    \subfloat{\includegraphics[width=0.1\textwidth]{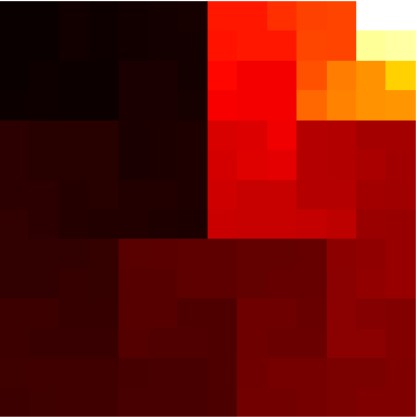}} \\

    \textbf{(13,13)} & \subfloat{\includegraphics[width=0.1\textwidth]{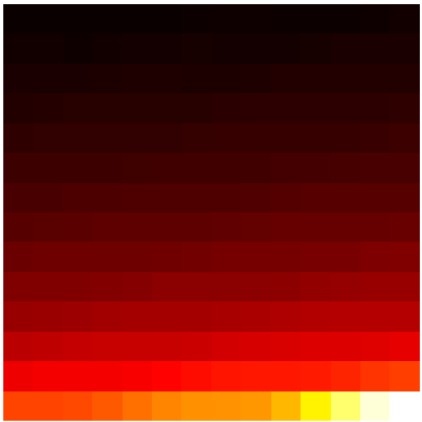}} &
    \subfloat{\includegraphics[width=0.1\textwidth]{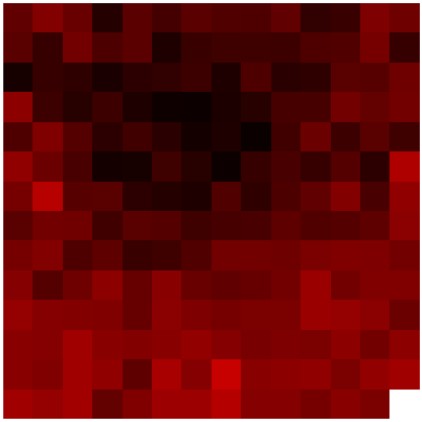}} &
    \subfloat{\includegraphics[width=0.1\textwidth]{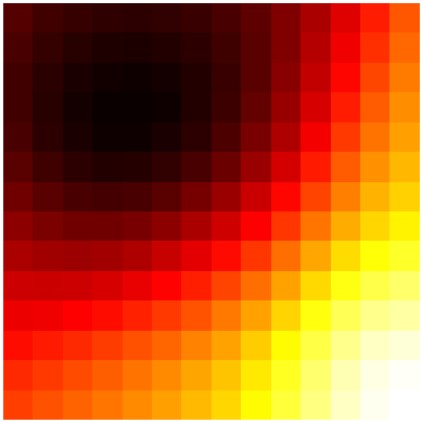}} &
    \subfloat{\includegraphics[width=0.1\textwidth]{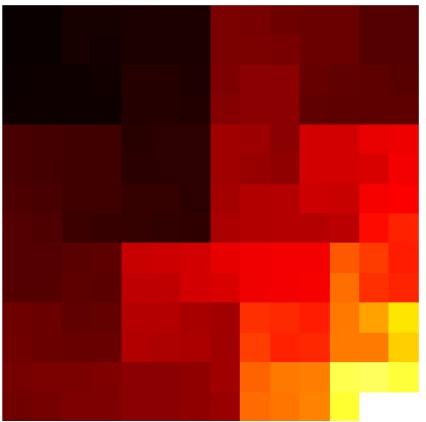}} \\

    \textbf{(0,7)} & \subfloat{\includegraphics[width=0.1\textwidth]{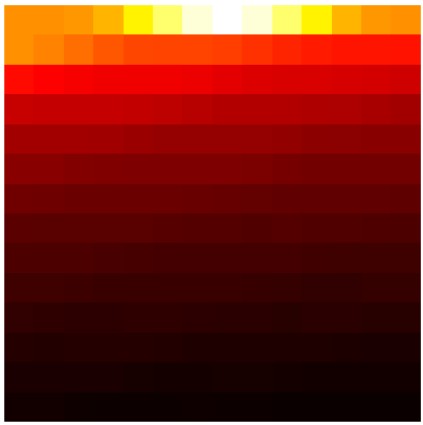}} &
    \subfloat{\includegraphics[width=0.1\textwidth]{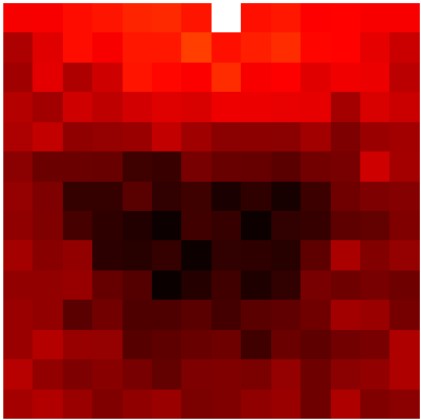}} &
    \subfloat{\includegraphics[width=0.1\textwidth]{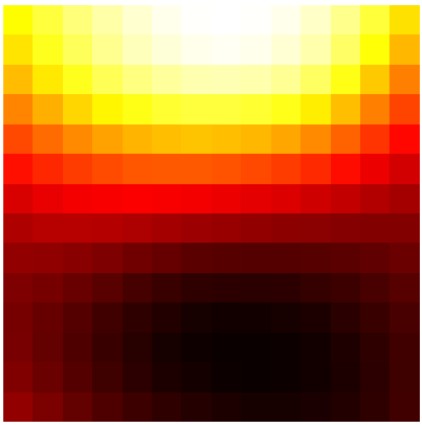}} &
    \subfloat{\includegraphics[width=0.1\textwidth]{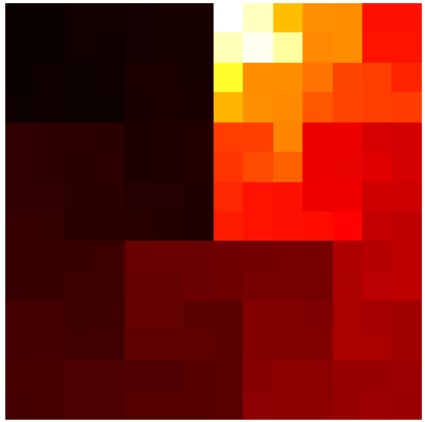}} \\

    \textbf{(7,0)} & \subfloat{\includegraphics[width=0.1\textwidth]{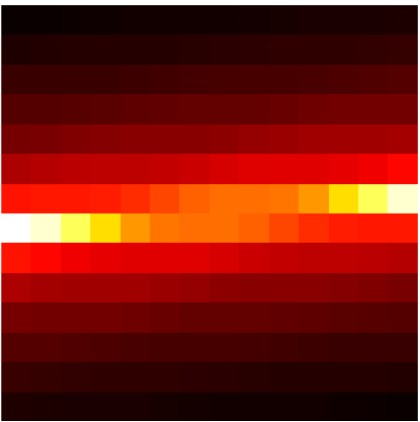}} &
    \subfloat{\includegraphics[width=0.1\textwidth]{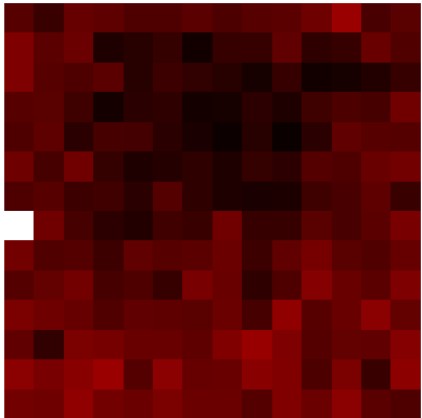}} &
    \subfloat{\includegraphics[width=0.1\textwidth]{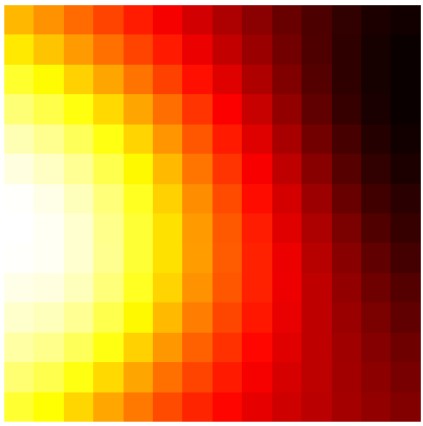}} &
    \subfloat{\includegraphics[width=0.1\textwidth]{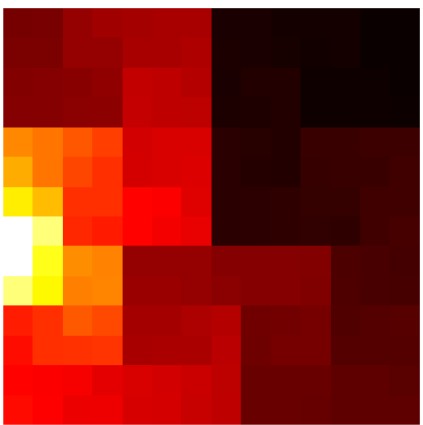}} \\

    \textbf{(7,13)} & \subfloat{\includegraphics[width=0.1\textwidth]{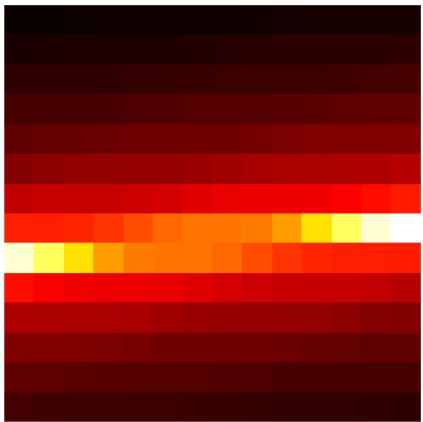}} &
    \subfloat{\includegraphics[width=0.1\textwidth]{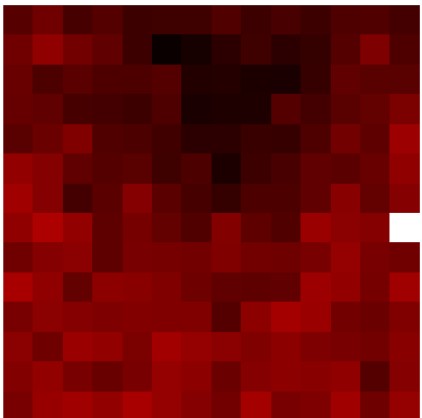}} &
    \subfloat{\includegraphics[width=0.1\textwidth]{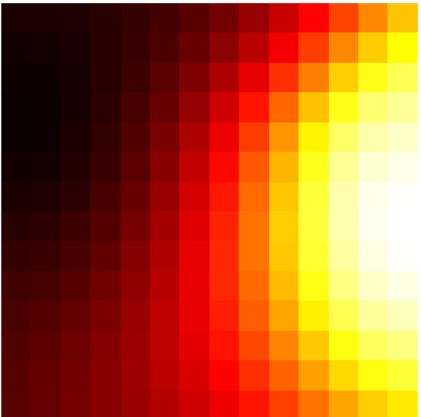}} &
    \subfloat{\includegraphics[width=0.1\textwidth]{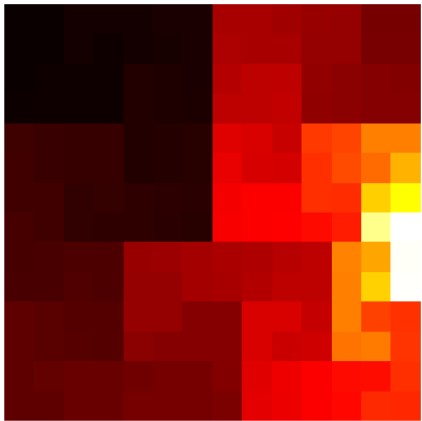}} \\

    \textbf{(13,7)} & \subfloat{\includegraphics[width=0.1\textwidth]{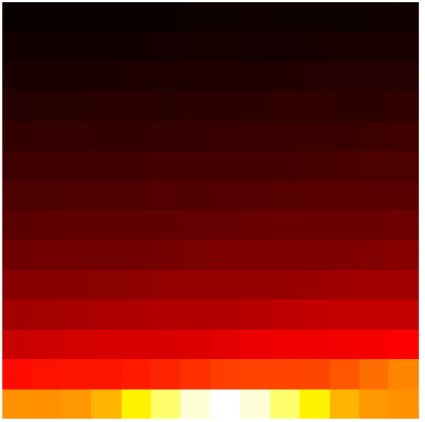}} &
    \subfloat{\includegraphics[width=0.1\textwidth]{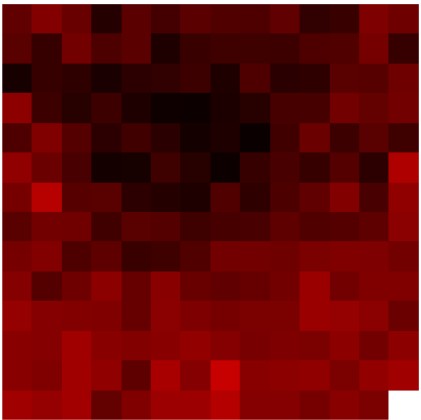}} &
    \subfloat{\includegraphics[width=0.1\textwidth]{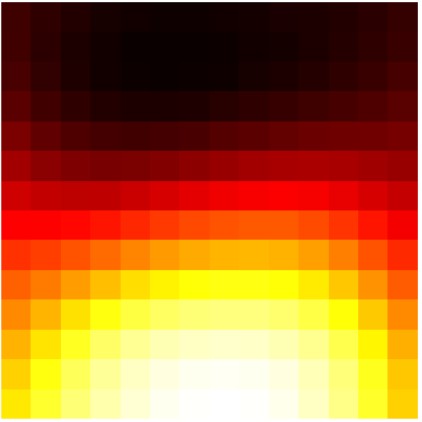}} &
    \subfloat{\includegraphics[width=0.1\textwidth]{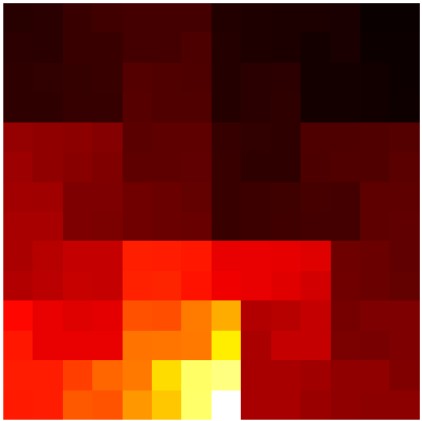}} \\

    \textbf{(7,7)} & \subfloat{\includegraphics[width=0.1\textwidth]{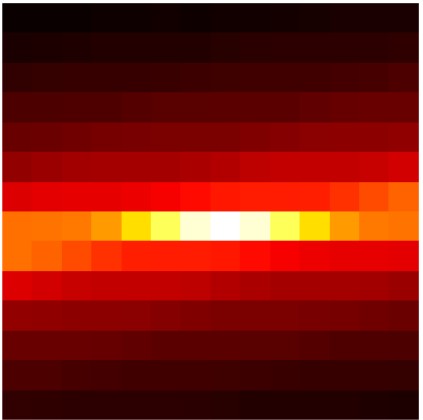}} &
    \subfloat{\includegraphics[width=0.1\textwidth]{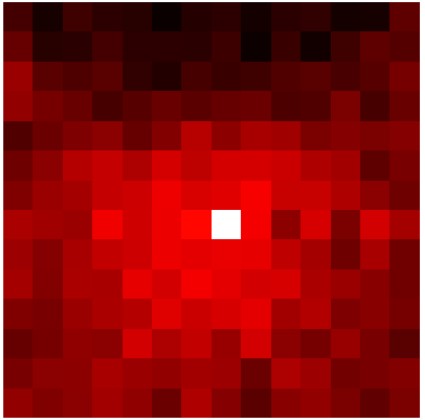}} &
    \subfloat{\includegraphics[width=0.1\textwidth]{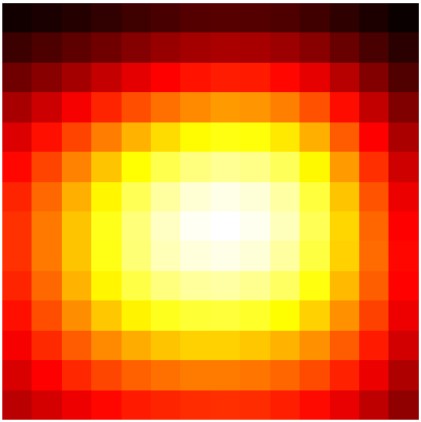}} &
    \subfloat{\includegraphics[width=0.1\textwidth]{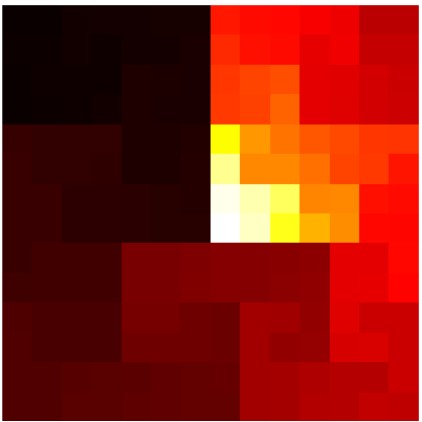}} \\
    \end{tabular}
    \caption{Cosine Similarity Map Comparison among varios PEs on different positions}
    \label{fig:9x4grid}
\end{figure}

Since, showing exactly how the properties in Table~1, retains in real-life vision tasks is quite hard and the lone contribution of PEs to the performance of ViTs are not huge (generally 4-6\%). That's why generating an edge case, where any property may fail/pass, is barely possible.
Instead, we can visually verify the implication of PESI metrics in the plotted correlation map of Positional encoding. In figure \ref{fig:9x4grid}, it has shown the correlation map with different location of patch, on different embedding policy.

From instance, according to Table~6, LFF(fourier) have highest Directed Monotonicity, \(M_D\), as any viewer can see, LFF has the smoothest descending correlation in any particular direction. But, LOOPE has the highest score in Undirected Monotonicity.

\end{document}